\documentclass[preprint,sort,compress,12pt]{elsarticle}
\usepackage{graphicx}
\usepackage{amsmath}
\usepackage{amssymb}
\usepackage{mathrsfs}
\usepackage{array}
\usepackage{upgreek}
\usepackage{float}
\usepackage{booktabs}
\usepackage{adjustbox}
\usepackage{makecell}
\usepackage[margin=1in]{geometry}
\usepackage[small,compact]{titlesec}
\usepackage[normal,bf,labelsep=colon,skip=2pt]{caption}
\usepackage[titles]{tocloft}
\usepackage{bold-extra}
\usepackage{hyperref}
\usepackage[ruled,vlined]{algorithm2e}
\usepackage{caption}
\usepackage{multirow}
\usepackage{tabularx}
\usepackage{diagbox}
\usepackage{subcaption}
\usepackage{bm}
\usepackage{enumitem}
\usepackage{algpseudocode}
\usepackage{amsfonts} 

\DeclareMathOperator{\KL}{KL}

\newcommand{\Tau}{\mathrm{T}}

\DeclareMathOperator*{\argmin}{arg\,min}

\usepackage{xcolor}
\hypersetup{
    colorlinks,
    linkcolor={red!50!black},
    citecolor={blue!50!black},
    urlcolor={blue!80!black}
}

\begin{document}
\begin{frontmatter}
\title{Synergizing Transport-Based Generative Models and Latent Geometry for Stochastic Closure Modeling}

\author{Xinghao Dong}
\author{Huchen Yang}
\author{Jin-Long Wu\corref{cor1}} \ead{jinlong.wu@wisc.edu} 
\cortext[cor1]{Corresponding author}

\address{Department of Mechanical Engineering, University of Wisconsin–Madison, Madison, WI 53706}

\begin{abstract}
Diffusion models recently developed for generative AI tasks can produce high-quality samples while still maintaining diversity among samples to promote mode coverage, providing a promising path for learning stochastic closure models. Compared to other types of generative AI models, such as GANs and VAEs, the sampling speed is known as a key disadvantage of diffusion models. By systematically comparing transport-based generative models on a numerical example of 2D Kolmogorov flows, we show that flow matching in a lower-dimensional latent space is suited for fast sampling of stochastic closure models, enabling single-step sampling that is up to two orders of magnitude faster than iterative diffusion-based approaches. To control the latent space distortion and thus ensure the physical fidelity of the sampled closure term, we compare the implicit regularization offered by a joint training scheme against two explicit regularizers: metric-preserving (MP) and geometry-aware (GA) constraints. Besides offering a faster sampling speed, both explicitly and implicitly regularized latent spaces inherit the key topological information from the lower-dimensional manifold of the original complex dynamical system, which enables the learning of stochastic closure models without demanding a huge amount of training data. 
\end{abstract}


\end{frontmatter}

\section{Introduction}
Complex dynamical systems, such as turbulent flows~\cite{pope2001turbulent} or solid mechanics~\cite{hughes1998variational} in engineering applications and physical processes in the Earth system~\cite{schneider2017earth}, are often featured by interactions across vast and continuous scales of space and time. The computational cost of fully resolving every scale in a Direct Numerical Simulation (DNS) is often prohibitive~\cite{moin1997tackling} for real-world science and engineering problems, and practical numerical simulations need to rely on closure models to approximate the impact of unresolved, small-scale dynamics on the numerically resolved coarse-grained variables. Most existing methods, e.g., RANS or LES closures for modeling turbulence, rely on a deterministic assumption, which only approximately holds if the unresolved scales achieve equilibrium in a time scale much faster than the one that those resolved scales evolve with. However, such a separation between resolved and unresolved scales may not exist for certain problems where the unresolved scales are far from equilibrium, motivating recent studies of going beyond the deterministic closures and exploring stochastic modeling approaches~\cite{palmer2010stochastic}.

Stochastic modeling has been explored for complex dynamical systems such as turbulence, for several decades~\cite {kraichnan1966dispersion,monin1971statistical}, leading to the development of stochastic models for some complex features of turbulent flows, e.g., intermittency~\cite{kraichnan1990models} and back scattering~\cite{mason1992stochastic}. Starting around the 
millennium, a substantial amount of research about stochastic models was explored for geophysical flows~\cite{majda1999models,majda2001mathematical, majda2003systematic}, with an excellent review of stochastic modeling for weather and climate presented by~\cite{palmer2019stochastic}. In the meantime, stochastic modeling techniques such as random matrices were also explored in solid mechanics~\cite{soize2005random} to account for the model uncertainties. More recently, mesoscale stochastic approaches were explored in the modeling of many complex systems, such as metallic foams~\cite{seif2025stochastic} and cellular interactions~\cite{dinapoli2021mesoscale}. From a broader perspective, stochasticity naturally shows up in reduced-order modeling techniques such as Mori-Zwanzig formalism, which demonstrates that when fast-evolving variables are integrated out of a system, their influence on the slow variables manifests as both a modified deterministic force and essential memory (non-Markovian) and stochastic noise terms~\cite{zwanzig1961memory,mori1965transport}. In practice, stochastic parameterizations have been shown to sharpen mean predictions, restore physical multi-modal variability, and reproduce the heavy-tailed statistics of extreme events across a wide range of applications~\cite{franzke2015stochastic,zwanzig2001nonequilibrium,palmer2019stochastic, ajayamohan2013realistic, wu2020enforcing, chen2023stochastic, xiao2016quantifying}. However, developing and calibrating stochastic closures present their own significant challenges~\cite{kondrashov2015data, gupta2023generalized, feng2024validity, bhouri2022history}, which often pose a more complicated model structure than classical deterministic closures and thus underscore the need for both a larger amount of data and a more sophisticated calibration procedure.


This need can be addressed by the growing field of scientific machine learning (SciML), which seeks to augment or replace traditional scientific modeling pipelines with machine learning techniques \cite{baker2019workshop, bergen2019machine, wang2023scientific, carleo2019machine}. Broadly, SciML efforts in dynamical systems modeling follow two main thrusts. The first thrust aims to create data-driven surrogates that approximate the system's evolution from data, effectively replacing traditional physics-based models, e.g., via system identification~\cite{brunton2016discovering, champion2019data, chen2023ceboosting, gao2025sparse} or operator learning~\cite{lu2019deeponet,li2020fourier,chen2025modeling}. The second thrust~\cite{wang2017physics,wu2018physics,kashinath2021physics,wu2024learning,dong2025data,yang2025bayesian, yang2025active}, as the focus of this work, uses machine-learning-based models not to replace the traditional physics-based solver but to augment it. This is the goal of data-driven closure modeling, which retains the well-established physical solver for the resolved scales and uses a learned model for the contributions from unresolved ones. It is worth noting that many research works (e.g.,~\cite{wang2017physics,wu2018physics}) in the second thrust adopted a deterministic form of the machine-learning-based models, while the recent advances in generative AI techniques opened up the possibilities of systematically constructing and calibrating data-driven stochastic closure models~\cite{dong2025data}.

Among the recent developments of generative AI techniques, three key paradigms, all united under a general transport-based framework, have emerged as compelling solutions:
\begin{itemize}
    \item \textbf{Score-based Diffusion Models} transform data into a simple prior distribution (typically Gaussian noise) through a fixed forward SDE and then learn to reverse this process with a learned score function. This approach has been successfully applied to stochastic closure modeling~\cite{dong2025data} and excels at capturing rich, non-Gaussian posteriors, but their highly curved transport paths necessitate slow, iterative sampling with hundreds of solver steps to maintain fidelity \cite{ho2020denoising, song2020score, song2019generative}. The extension to conditional diffusion models has been explored by various computational mechanics problems~\cite{gao2024bayesian,du2024conditional,dasgupta2025conditional,jacobsen2025cocogen,dong2025data,dong2025stochastic}.
    \item \textbf{Flow Matching} replaces the stochastic noising path with a simpler, often linear, interpolation between noise and data. It then learns a deterministic ODE velocity field to transport samples along these straight paths. This formulation dramatically simplifies the transport, enabling high-quality generation in a single step and admitting exact likelihood computation, though potentially at the cost of reduced intrinsic randomness \cite{papamakarios2021normalizing, lipman2022flow, liu2022flow, tong2023conditional}.
    \item \textbf{Stochastic Interpolants} provide a unifying perspective, defining a transport process that explicitly interpolates between two distributions while allowing for the injection of time-dependent noise. This framework retains the efficient, straight paths of FM while restoring the stochastic expressiveness and flexibility of diffusion models \cite{albergo2023stochastic, chen2024probabilistic, albergo2022building}.
\end{itemize}

These compelling solutions of generative AI techniques, originally developed for standard machine learning tasks such as image/video generation, motivate a central question of this work: for stochastic closure modeling, where rapid and repeated sampling is essential, which of these paradigms best navigates the critical trade-off between sampling speed, sample quality, and uncertainty representation? In addition, since the iterative cost of any transport-based sampler scales with the dimensionality of the space to perform sampling, a complementary strategy for addressing their computational bottlenecks involves shifting expensive operations to lower-dimensional latent spaces. Latent space generative models \cite{rombach2022high, vahdat2021score, luo2023refusion, dao2023flow} offer a promising enhancement through a two-stage pipeline: autoencoders compress high-dimensional data into compact representations, then generative processes operate within this reduced space. For online closure modeling, where a new sample is required at each time step of a physics-based simulation, this can accelerate the total simulation time by orders of magnitude~\cite{dong2025stochastic}.

The success of sampling in a latent space, however, is entirely contingent on the quality and structure of the learned latent representation. A standard autoencoder, trained solely to minimize reconstruction error, has no incentive to preserve the geometric or statistical structure of the original data manifold, which can potentially force the generative model to learn much more complicated (or even ill-posed) dynamics in the latent space, leading to unsatisfactory training and inaccurate sampling performances. 

To overcome this challenge, the latent space must be explicitly structured. One approach is implicit regularization via end-to-end joint training, which forces the autoencoder to learn a representation aligned with the generative task, outperforming conventional two-phase methods~\cite{dong2025stochastic}. However, this offers no direct control over the resulting geometry. A more principled strategy, which has gained traction in the broader machine learning community, is to employ explicit regularizers during autoencoder training. These methods enforce desired inductive biases, such as spatial equivariance \cite{kouzelis2025eqvaeequivarianceregularizedlatent}, multiscale consistency via wavelet-based penalties \cite{sigillo2025latentwaveletdiffusionenabling}, or geometric alignment through contrastive losses \cite{Zhou_2025_CVPR, sun2024geometry}. In this work, we study two types of regularizer: geometry-aware (GA) regularization and metric-preserving (MP) constraints. These techniques craft a latent space that mirrors the geometric and topological features of the original data, directly improving the efficiency and accuracy of generative models for stochastic closure applications.

To summarize, this paper makes the following key contributions:

\begin{itemize}
    \item We perform the first systematic comparison of diffusion, flow matching, and stochastic interpolant paradigms for stochastic closures. We show that flow-based methods achieve superior sampling speed via straighter transport paths, enabling order-of-magnitude reductions in the number of integration steps with minimal error increase.
    \item We demonstrate that naive, reconstruction-only autoencoders introduce significant geometric distortions that scatter conditional distributions. We show that both implicit (joint training) and our proposed explicit (MP, GA) regularization strategies mitigate these issues, yielding structured latent spaces with quantifiable reductions in distortion and improved sample fidelity.
    \item We show that the resulting regularized latent generators integrate seamlessly into physics-based solvers, delivering efficient uncertainty quantification that reproduces full-system statistics while accelerating overall simulation time.
\end{itemize}

\section{Methodology}\label{sec: methodology}
We consider spatiotemporal dynamical systems, such as those describing turbulent flows and weather patterns, governed by the full-order equations:
\begin{equation}
    \label{eqn:true_system}
    \frac{\partial \mathbf{v}}{\partial t} = \mathcal{M}(\mathbf{v}),
\end{equation}
where $\mathbf{v} \in \mathcal{V}$ denotes the state encompassing all scales and $\mathcal{M}$ is the nonlinear dynamical operator. The vast range of scales often renders full resolution computationally intractable, necessitating reduced-order formulations that evolve only the resolved state, $\mathbf{V} = \mathcal{P}(\mathbf{v})$:
\begin{equation}
    \label{eqn:corrected_system}
    \frac{\partial \mathbf{V}}{\partial t} = \overline{\mathcal{M}}(\mathbf{V}) + \mathcal{C}(\mathbf{V}).
\end{equation}
Here, $\mathcal{P}$ is a projection operator (e.g., a low-pass filter or encoder mapping) and $\overline{\mathcal{M}}$ is the projected dynamical operator. The closure operator, $\mathcal{C}$, is necessary because the projection $\mathcal{P}$ does not commute with the nonlinear dynamics $\mathcal{M}$. This non-commutation means that the evolution of the resolved state $\mathbf{V}$ depends on interactions with the unresolved scales. The closure term, $\mathbf{U} = \mathcal{C}(\mathbf{V})$, models the net effect of these missing physical interactions—such as energy backscatter and turbulent dissipation—and is essential for restoring fidelity to the reduced system.

Traditional approaches that parameterize $\mathbf{U}$ with simple deterministic or stochastic functions often fail to capture its complex, non-Gaussian, and history-dependent nature. A more powerful, data-driven approach is to treat the closure term as a random object and learn its full conditional distribution, $p(\mathbf{U} | \mathbf{V})$, from high-fidelity data. To achieve a rich, stochastic formulation, we can conceptualize the closure term $\mathbf{U}$ itself as a stochastic field whose dynamics evolve according to:
\begin{equation}
    \label{eqn:closure_spde}
    \frac{\partial \mathbf{U}}{\partial t} = \mathcal{H}(\mathbf{U}; \mathbf{V}) + \boldsymbol{\xi},
\end{equation}
where $\mathcal{H}$ encompasses the operators governing the evolution of $\mathbf{U}$ conditioned on the resolved state $\mathbf{V}$, and $\boldsymbol{\xi}$ represents a stochastic forcing term. Rather than explicitly learning the complex Stochastic Partial Differential Equation (SPDE) in Eq.~\eqref{eqn:closure_spde}, we adopt a transport-based generative modeling approach to directly characterize the stationary conditional distribution $p(\mathbf{U} | \mathbf{V})$ that results from these dynamics.

This approach was first pioneered using latent-space score-based diffusion models, which demonstrated its viability for this task \cite{dong2025data, dong2025stochastic}. However, these foundational studies also highlighted the need for faster sampling paradigms to be practical in online simulations and for more robust latent space representations to ensure physical fidelity. To address these challenges, we develop a comprehensive framework in this section. We begin in Section~\ref{ssec:generative_models} by systematically comparing three transport-based generative paradigms—diffusion models, flow matching, and stochastic interpolants—to identify the optimal balance of speed and accuracy. Then, in Section~\ref{ssec:Latent_Models}, we introduce and evaluate several strategies for crafting geometrically structured latent spaces to enhance the physical consistency of the generated closures.

\subsection{Transport-based Latent Generative Models for Stochastic Closures}\label{ssec:generative_models}
While score-based diffusion models, flow matching, and stochastic interpolants all operate by learning a map from a simple prior distribution to a complex target distribution, they differ fundamentally in their transport mechanisms. These differences in how they move probability mass result in distinct sampling procedures, computational demands, and training objectives. The paradigms span a spectrum from stochastic to deterministic transport and from highly curved to linear sampling paths.

\subsubsection{Score-based Diffusion Models}\label{sssec: diffusion_models}
Score-based diffusion models are a class of generative models that produce samples by reversing a predefined noise-injection process. The framework consists of two parts: a fixed forward process that gradually perturbs data into noise via a Stochastic Differential Equation (SDE), and a learned reverse process that transforms noise back into data by solving a corresponding reverse-time SDE.

The forward process maps a data sample $\mathbf{x}_0 \sim p_{\text{data}}(\mathbf{x})$ to a noise vector over a continuous time interval $\tau \in [0, \Tau]$. A common choice is the Variance-Exploding (VE) SDE~\cite{song2020score}:
\begin{equation}\label{eqn:VESDE}
    \mathrm{d}\mathbf{x} = \sigma^\tau \mathrm{d}\mathbf{W},
\end{equation}
where $\sigma > 1$ is a hyperparameter and $\mathbf{W}$ is a standard Wiener process. This forward process is a special case of the Ornstein–Uhlenbeck process and defines a Markov chain with an analytical transition kernel:
\begin{equation}\label{eqn: transitionkernel}
p(\mathbf{x}_\tau \mid \mathbf{x}_0) = \mathcal{N}\left(\mathbf{x}_\tau \mid \boldsymbol{\mu}(\mathbf{x}_0, \tau), \mathbf{\Sigma}(\tau)\right),
\end{equation}
where
\begin{equation}
\boldsymbol{\mu}(\mathbf{x}_0, \tau) = \mathbf{x}_0, \quad \mathbf{\Sigma}(\tau) = \frac{1}{2 \log \sigma}\left(\sigma^{2\tau}-1\right) I.
\end{equation}
As $\tau \to \Tau$, the distribution $p(\mathbf{x}_\tau)$ approaches an isotropic Gaussian independent of the original data, from which we can easily sample:
\begin{equation}\label{eqn: terminal_dist}
p(\mathbf{x}_\Tau) = \int p(\mathbf{x}_0)p(\mathbf{x}_\Tau \mid \mathbf{x}_0) \mathrm{d}\mathbf{x}_0 \approx \mathcal{N}\left(0, \frac{1}{2 \log \sigma}\left(\sigma^{2\Tau}-1\right) I \right).
\end{equation}

A known result from stochastic calculus states that this forward process has a corresponding reverse-time SDE, which allows us to reverse the noising process to generate data \cite{anderson1982reverse}:
\begin{equation}\label{eqn:reverse_SDE}
    \mathrm{d}\mathbf{x} = -\sigma^{2\tau} \nabla_{\mathbf{x}_\tau} \log p(\mathbf{x}_\tau) \mathrm{d}\tau + \sigma^\tau \mathrm{d}\overline{\mathbf{W}},
\end{equation}
where $\mathrm{d}\overline{\mathbf{W}}$ is a Wiener process running backward in time. Critically, solving this SDE requires the score function, $\nabla_{\mathbf{x}_\tau} \log p(\mathbf{x}_\tau)$, of the marginal noisy data distribution $p(\mathbf{x}_\tau)$, which is intractable.

The central task is therefore to learn a neural network, $\mathbf{s}_\theta(\tau, \mathbf{x}_\tau)$, to approximate this score. This is achieved via denoising score matching, where the network is trained to predict the score of the analytically known conditional distribution $p(\mathbf{x}_\tau | \mathbf{x}_0)$. The training objective minimizes the weighted squared error between the network's output and the conditional score:
\begin{equation}
    \label{eqn:dsm_loss}
    \min_\theta \mathbb{E}_{\tau, \mathbf{x}_0, \mathbf{x}_\tau} \left[ \lambda(\tau) \left\| \mathbf{s}_\theta(\tau, \mathbf{x}_\tau) - \nabla_{\mathbf{x}_\tau} \log p(\mathbf{x}_\tau | \mathbf{x}_0) \right\|_2^2 \right],
\end{equation}
where $\lambda(\tau)$ is a positive weighting function that balances the loss across different noise levels to improve training stability. For many diffusion setups, this simplifies to setting the weighting equal to the variance of the added noise, i.e. $\lambda(\tau) = \mathbf{\Sigma}(\tau)$. The conditional score is simply $-(\mathbf{x}_\tau - \mathbf{x}_0) / \Sigma(\tau)$. This objective is tractable as it relies only on samples from the forward process and has been shown to be equivalent to matching the true marginal score~\cite{vincent2011connection, song2020sliced}.

Once trained, $\mathbf{s}_\theta$ is used as a plug-in estimator for the true score in Eq.~\eqref{eqn:reverse_SDE}. New samples are generated by starting with $\mathbf{x}_\Tau$ drawn from the Gaussian prior and numerically integrating the SDE backward in time, for instance with an Euler-Maruyama solver. This iterative sampling procedure is powerful but computationally expensive, often requiring hundreds of steps to maintain fidelity due to the curved nature of the diffusion paths. 

For conditional modeling of $p(\mathbf{x} | \mathbf{y})$, the framework is extended by modifying the score network to accept the condition $\mathbf{y}$ as an additional input. During each training step, a data pair $(\mathbf{x}_0, \mathbf{y})$ is sampled, and the forward noising process is applied only to $\mathbf{x}_0$. The score network $\mathbf{s}_\theta(\tau, \mathbf{x}_\tau, \mathbf{y})$ then uses both the noisy data and the clean condition to predict the score, with the training objective remaining analogous to the unconditional case, except that initial samples are drawn from the joint distribution $p(\mathbf{x}, \mathbf{y})$. In addition, the forward diffusion process acts exclusively on the target variable $\mathbf{x}$, meaning the perturbing kernel $p(\mathbf{x}_\tau | \mathbf{x}_0)$ is conditionally independent of any input condition $\mathbf{y}$. Thus, the training loss becomes:
\begin{equation}
\theta^* = \argmin_\theta \mathbb{E}_{\tau, (\mathbf{x}_0, \mathbf{y}), \mathbf{x}_\tau} \left[ \lambda(\tau)\left\| \mathbf{s}_\theta(\tau, \mathbf{x}_\tau, \mathbf{y}) - \nabla_{\mathbf{x}_\tau} \log p(\mathbf{x}_\tau | \mathbf{x}_0) \right\|^2 \right],
\end{equation}
where expectations are taken over $\tau \sim \mathcal{U}[0, T]$, $(\mathbf{x}_0, \mathbf{y}) \sim p(\mathbf{x}, \mathbf{y})$, and $\mathbf{x}_\tau \sim p(\mathbf{x}_\tau \mid \mathbf{x}_0)$.

\subsubsection{Flow Matching}
Flow Matching (FM) is a paradigm for training continuous-time generative models that avoids the complexities of SDEs by learning a deterministic velocity field $\mathbf{v}_\theta$. This learned field defines an Ordinary Differential Equation (ODE) that transports samples from a simple prior distribution to the target data distribution.

The core idea is to define a probability path $p_\tau(\mathbf{x})$ that transitions from a prior $p_0(\mathbf{x}) \approx \mathcal{N}(\mathbf{0}, \mathbf{I})$ at $\tau=0$ to the data distribution $p_1(\mathbf{x}) = p_{\text{data}}(\mathbf{x})$ at $\tau=1$. This path is generated by a true, underlying marginal velocity field $\mathbf{v}(\tau, \mathbf{x})$. Ideally, one would train the neural network $\mathbf{v}_\theta$ by directly minimizing the discrepancy between it and this true field:
\begin{equation}
    \mathcal{L}_\text{FM}(\theta) = \mathbb{E}_{\tau \sim \mathcal{U}[0, 1], \, \mathbf{x}_\tau \sim p_\tau(\mathbf{x})} \left\| \mathbf{v}_\theta(\tau, \mathbf{x}_\tau) - \mathbf{v}(\tau, \mathbf{x}_\tau) \right\|_2^2.
\end{equation}
However, this objective is intractable because both the marginal probability path $p_\tau(\mathbf{x})$ and its velocity field $\mathbf{v}(\tau, \mathbf{x})$ are unknown.

Conditional Flow Matching (CFM) resolves this issue with a key insight: instead of working with intractable marginal paths, we can define a simple, tractable conditional path and velocity field, and train the model to match those instead \cite{lipman2022flow, tong2023conditional}. Specifically, we define a path conditioned on a sample from the prior, $\mathbf{x}_0 \sim p_0(\mathbf{x})$, and a sample from the data, $\mathbf{x}_1 \sim p_{\text{data}}(\mathbf{x})$. A common and effective choice is a linear interpolation path:
\begin{equation}\label{eqn:linear_path}
    p(\mathbf{x}_\tau \mid \mathbf{x}_0, \mathbf{x}_1) = \delta\left(\mathbf{x}_\tau - \left[(1 - \tau) \mathbf{x}_0 + \tau \mathbf{x}_1\right]\right),
\end{equation}
which has a simple, constant conditional velocity:
\begin{equation}\label{eqn:conditional_velocity}
    \mathbf{v}(\tau, \mathbf{x}_\tau \mid \mathbf{x}_0, \mathbf{x}_1) = \mathbf{x}_1 - \mathbf{x}_0.
\end{equation}
The central theorem of CFM shows that a loss defined on these simple conditional quantities has the same expected gradient as the intractable marginal loss. This leads to a practical and efficient training objective:
\begin{equation}\label{eqn:CFM_Loss}
    \min_\theta \mathbb{E}_{\tau, \mathbf{x}_0, \mathbf{x}_1} \left\| \mathbf{v}_\theta(\tau, (1-\tau)\mathbf{x}_0 + \tau\mathbf{x}_1) - (\mathbf{x}_1 - \mathbf{x}_0) \right\|_2^2.
\end{equation}
This objective is a simple regression problem that does not require simulating an ODE during training.

Once the velocity field $\mathbf{v}_\theta$ is trained, new samples are generated by solving the initial value problem for the generation ODE, starting from a random sample $\mathbf{x}_0 \sim p_0(\mathbf{x})$:
\begin{equation}\label{eqn:generation_ode}
    \frac{d \mathbf{x}}{d \tau} = \mathbf{v}_\theta(\tau, \mathbf{x}), \quad \text{for } \tau \in [0, 1].
\end{equation}
This ODE can be solved with standard numerical integrators, such as the Euler method. Because the training encourages nearly straight transport paths, FM models are highly efficient at inference, often requiring only 10--100 steps for high-quality generation—a significant speed-up over typical diffusion models.

For conditional modeling of $p(\mathbf{x} | \mathbf{y})$, the velocity network is simply modified to accept the condition $\mathbf{y}$ as an additional input, $\mathbf{v}_\theta(\tau, \mathbf{x}_\tau, \mathbf{y})$. The training objective in Eq.~\eqref{eqn:CFM_Loss} is adapted by sampling from the joint data distribution $(\mathbf{x}_1, \mathbf{y}) \sim p(\mathbf{x}, \mathbf{y})$.

\subsubsection{Stochastic Interpolants}
Stochastic Interpolants (SI) offer a unifying and highly flexible paradigm for generative modeling that generalizes both diffusion models and flow matching \cite{albergo2023stochastic}. The core idea is to explicitly define a stochastic path, or interpolant, that connects any arbitrary source distribution $p_0(\mathbf{x}_0)$ to the target data distribution $p_1(\mathbf{x}_1)$, and then learn the drift of the SDE that generates this path.

Formally, the interpolant path between a pair of samples $(\mathbf{x}_0, \mathbf{x}_1)$ is defined as:
\begin{equation}\label{eqn:SI_path}
    \mathbf{x}_\tau = \alpha_\tau \mathbf{x}_0 + \beta_\tau \mathbf{x}_1 + \sigma_\tau \mathbf{W}_\tau, \quad \tau \in [0, 1],
\end{equation}
where $\mathbf{W}_\tau$ is a standard Wiener process and the coefficients $(\alpha_\tau, \beta_\tau, \sigma_\tau)$ satisfy the boundary conditions $\mathbf{x}_{\tau=0} = \mathbf{x}_0$ and $\mathbf{x}_{\tau=1} = \mathbf{x}_1$. The SI framework provides a simple and efficient training objective based on this path. We first define a path velocity $\mathbf{r}_\tau$, which includes a drift term that arises from the time-varying noise schedule:
\begin{equation}
    \mathbf{r}_\tau = \dot{\alpha}_\tau \mathbf{x}_0 + \dot{\beta}_\tau \mathbf{x}_1 + \dot{\sigma}_\tau \mathbf{W}_\tau,
\end{equation}
where the dot denotes a time derivative. The neural network $\mathbf{b}_\theta(\tau, \mathbf{x}_\tau)$, which approximates the drift of the forward SDE, is trained via a simple regression objective to predict this path velocity:
\begin{equation}
    \min_\theta \mathbb{E}_{\tau, \mathbf{x}_0, \mathbf{x}_1, \mathbf{W}_\tau} \left\| \mathbf{b}_\theta(\tau, \mathbf{x}_\tau) - \mathbf{r}_\tau \right\|_2^2.
\end{equation}
A common choice that encourages straight mean paths is the \textbf{linear interpolant}, where $\alpha_\tau = 1-\tau$ and $\beta_\tau = \tau$. In this case, the deterministic part of the path velocity simplifies to the constant vector $\mathbf{x}_1 - \mathbf{x}_0$. The full path velocity, however, retains its stochastic component, which depends on the chosen noise schedule. For example, with a simple noise schedule of $\sigma_\tau = 1-\tau$, we have $\dot{\sigma}_\tau = -1$, and the full path velocity becomes:
\begin{equation}
    \mathbf{r}_\tau = \mathbf{x}_1 - \mathbf{x}_0 - \mathbf{W}_\tau.
\end{equation}
This objective mirrors Conditional Flow Matching but crucially incorporates the stochastic drift term, allowing for tunable noise injection.

Once trained, new samples are generated by numerically integrating the learned forward SDE, starting with a sample from the prior:
\begin{equation}\label{eqn:SI_SDE}
    \mathrm{d}\mathbf{x}_\tau = \mathbf{b}_\theta(\tau, \mathbf{x}_\tau) \mathrm{d}\tau + \sigma_\tau \mathrm{d}\mathbf{W}_\tau, \quad \mathbf{x}_0 \sim p_0(\mathbf{x}),
\end{equation}
For conditional modeling of $p(\mathbf{x} | \mathbf{y})$, the drift network is modified to $\mathbf{b}_\theta(\tau, \mathbf{x}_\tau, \mathbf{y})$ and trained on samples from the joint distribution $p(\mathbf{x}_1, \mathbf{y})$.

A key advantage of the SI framework is its flexibility in choosing the source distribution $p_0$. While diffusion and standard FM typically use a fixed Gaussian prior, SI can directly interpolate between two arbitrary distributions. This is particularly powerful for closure modeling. Instead of starting from random noise, we can set the source distribution to be the conditional variable itself, $p_0(\mathbf{x}_0) = p(\mathbf{z}^\omega)$, and the target to be the closure, $p_1(\mathbf{x}_1) = p(\mathbf{z}^H | \mathbf{z}^\omega)$. The model then learns the direct, physically meaningful transport from the resolved state to the unresolved correction term.

\subsection{Crafting Structured Latent Spaces for Generative Closures}\label{ssec:Latent_Models}
Deploying transport-based generative models for stochastic closures is computationally challenging due to the high dimensionality of the discretized physical fields. The iterative sampling process can be prohibitively expensive when performed in the full state space. To mitigate this cost, we employ a latent space approach, using a convolutional autoencoder to learn a low-dimensional representation of the data. 

Given a physical field $\mathbf{U}$ (with the conditioning field $\mathbf{V}$ treated analogously), the encoder $\mathcal{E}_\mathbf{U}$ maps it to a compact latent vector:
\begin{equation}\label{eqn:encoder_U}
    \mathbf{z^U} = \mathcal{E}_\mathbf{U}(\mathbf{U}), \quad \text{where } \mathbf{U} \in \mathbb{R}^{d_\mathbf{U}} \text{ and } \mathbf{z^U} \in \mathbb{R}^{l_\mathbf{U}} \text{ with } l_\mathbf{U} \ll d_\mathbf{U}.
\end{equation}
The decoder $\mathcal{D}_\mathbf{U}$ then reconstructs an approximation of the original field from this latent vector, $\hat{\mathbf{U}} = \mathcal{D}_\mathbf{U}(\mathbf{z^U})$. In a conventional two-phase pipeline, the autoencoder parameters are optimized by solely minimizing the mean squared reconstruction error:
\begin{equation}\label{eqn:ae_loss}
    \mathcal{L}^\mathbf{U}_{\text{Recon}} = \mathbb{E}_{\mathbf{U} \sim p(\mathbf{U})} \| \mathbf{U} - \mathcal{D}_\mathbf{U}(\mathcal{E}_\mathbf{U}(\mathbf{U})) \|_2^2.
\end{equation}

However, a latent space optimized only for reconstruction quality can be arbitrarily distorted, as the loss in Eq.~\eqref{eqn:ae_loss} is agnostic to the manifold's geometric structure. This can scatter conditional distributions and complicate the transport paths for a subsequent generative model, degrading its performance. To overcome this limitation and craft a latent space that is well-suited for the generative task, we systematically compare two distinct strategies:
\begin{itemize}
    \item \textbf{End-to-end joint training}, where the autoencoder and generative model are optimized simultaneously, providing an \textit{implicit regularization} on the latent space.
    \item \textbf{Two-phase training with explicit regularization}, where the autoencoder is first pre-trained with an objective function that directly enforces specific geometric properties on the latent space.
\end{itemize}
These approaches aim to create more structured and informative latent spaces that simplify the generative task while maintaining high-fidelity reconstructions. The detailed model structures and training details of autoencoders can be found in~\ref{sec:training_details}.

\subsubsection{Implicit Regularization via Joint Training}\label{sssec:joint_training}
An alternative to the two-phase pipeline is to train the autoencoder and the generative model simultaneously. This end-to-end joint training serves as a powerful implicit regularizer. By receiving gradients from both the reconstruction and generative objectives, the autoencoder is forced to learn a latent space that is not only faithful to the original data but is also structured in a way that simplifies the generative transport task, often outperforming sequential training pipelines \cite{tompson2014joint, han2020joint}.

The training is guided by a multi-objective loss function, which is a weighted sum of three distinct terms:
\begin{equation}\label{eqn:joint_loss_full}
    \mathcal{L}_{\text{joint}} = \mathcal{L}_{\text{Recon}} + \lambda_{\text{Gen}} \mathcal{L}_{\text{Gen}} + \lambda_{\text{KL}} \mathcal{L}_{\text{KL}}.
\end{equation}
Each component addresses a different requirement of the learning process:

\begin{itemize}
    \item \textbf{The Reconstruction Loss ($\mathcal{L}_{\text{Recon}}$)} ensures that the autoencoder produces high-fidelity representations. It is a weighted mean squared error that often prioritizes the accuracy of the more complex or crucial field, which in this case is the closure term $\mathbf{U}$:
    \begin{equation}\label{eq:AE_loss_joint}
        \mathcal{L}_{\text{Recon}} = \mathbb{E}_{(\mathbf{U}, \mathbf{V})} \left[ \lambda_{\mathbf{U}} \| \mathbf{U} - \mathcal{D}_{\mathbf{U}}(\mathcal{E}_{\mathbf{U}}(\mathbf{U}))\|_2^2 + \lambda_{\mathbf{V}} \| \mathbf{V} - \mathcal{D}_{\mathbf{V}}(\mathcal{E}_{\mathbf{V}}(\mathbf{V}))\|_2^2 \right],
    \end{equation}
    where the expectation is over the data distribution, and typically $\lambda_{\mathbf{U}} > \lambda_{\mathbf{V}}$.

    \item \textbf{The Generative Loss ($\mathcal{L}_{\text{Gen}}$)} is the transport objective that trains the generative model. This corresponds to the score-matching, flow-matching, or drift-regression losses defined in Section~\ref{ssec:generative_models}.

    \item \textbf{The KL Regularization ($\mathcal{L}_{\text{KL}}$)} prevents latent collapse, a failure mode where the encoder maps all inputs to a small, uninformative region of the latent space \cite{lucas2019don}. This term encourages the aggregated distribution of encoded samples, $q(\mathbf{z^U})$, to match a simple prior, typically a standard Gaussian $p(\mathbf{z^U}) = \mathcal{N}(\mathbf{0}, \mathbf{I})$, thereby ensuring the latent space remains expressive:
    \begin{equation}\label{eq:KL_loss_joint}
        \mathcal{L}_{\text{KL}} = \KL\Bigl(q(\mathbf{z^U}) \,\| \, p(\mathbf{z^U})\Bigr).
    \end{equation}
\end{itemize}
The hyperparameters $\lambda_{\text{Gen}}$ and $\lambda_{\text{KL}}$ balance these competing objectives, and their tuning is critical for achieving a model that excels at both reconstruction and conditional generation.

\subsubsection{Explicit Regularization via a Two-Phase Strategy}\label{sssec:explicit_reg}
The second strategy for crafting a well-structured latent space is to employ explicit regularization within a stable, two-phase training pipeline. The core idea is to first pre-train the autoencoder with an objective function that directly enforces desired geometric properties on the latent space, before the generative model is trained. We introduce and evaluate two such regularizers: 

\begin{itemize}
    \item \textbf{Metric-Preserving (MP) Regularization:} This approach aims to make the encoder a \textit{local isometry}. It seeks to preserve the direct Euclidean distance between pairs of points, ensuring that the local "neighborhood" structure of the physical space is accurately mapped to the latent space. This is akin to "unrolling" the data manifold into a flat latent representation without locally stretching or tearing it.
    
    \item \textbf{Geometry-Aware (GA) Regularization:} This approach aims to preserve the more global, intrinsic \textit{manifold geometry}. Instead of using the straight-line Euclidean distance, it uses a pre-computed manifold distance (approximating the geodesic distance) between points. This captures the true "on-manifold" path length, preserving the larger-scale topological features of the data.
\end{itemize}

Both strategies are implemented by augmenting the standard reconstruction loss from Eq.~\eqref{eqn:ae_loss} with a structural loss term:
\begin{equation}
    \mathcal{L}_{\text{AE}} = \mathcal{L}^{\mathbf{U}}_{\text{Recon}} + \lambda_{\text{Struc}} \mathcal{L}^{\mathbf{U}}_{\text{Struc}},
\end{equation}
where $\lambda_{\text{Struc}}$ is a hyperparameter that balances reconstruction fidelity with geometric preservation. The structural loss, $\mathcal{L}_{\text{Struc}}$, penalizes the discrepancy between distances in the physical and latent spaces, with a focus on local neighborhoods:
\begin{equation}
    \mathcal{L}_{\text{Struc}} = \mathbb{E}_{\mathbf{U}_i, \mathbf{U}_j \sim p(\mathbf{U})} \left[ w(\mathbf{U}_i, \mathbf{U}_j) \left( \| \mathcal{E}_{\mathbf{U}}(\mathbf{U}_i) - \mathcal{E}_{\mathbf{U}}(\mathbf{U}_j)\|_2 - d(\mathbf{U}_i, \mathbf{U}_j) \right)^2 \right],
\end{equation}
where the weight $w(\mathbf{U}_i, \mathbf{U}_j) = e^{-\gamma d(\mathbf{U}_i, \mathbf{U}_j)}$ emphasizes local pairs. This general form is specialized by defining the distance metric $d(\cdot, \cdot)$ as either the Euclidean norm ($\| \cdot \|_2$) for MP regularization or the pre-computed manifold distance for GA regularization. This pre-training phase, applied to the autoencoders for both the closure and conditioning fields, produces latent representations that retain the intrinsic structure of the data, thereby simplifying the subsequent generative modeling task.

\section{Numerical Results}\label{sec: results}
\subsection{Numerical Setup}
We evaluate our generative closure framework on a two-dimensional stochastic Kolmogorov flow. The system is governed by the incompressible Navier-Stokes equations in vorticity form on a periodic domain $\Omega = (0, L)^2$ over the time interval $(0, T_{\text{phy}}]$:
\begin{equation}\label{eq:2dNS}
    \begin{aligned}
        \frac{\partial \omega(\mathbf{x}, t)}{\partial t} &= -\mathbf{u}(\mathbf{x}, t) \cdot \nabla\omega(\mathbf{x}, t) + f(\mathbf{x}) + \nu\nabla^2\omega(\mathbf{x}, t) + \beta \xi(\mathbf{x}, t), \\
        \nabla \cdot \mathbf{u}(\mathbf{x}, t) &= 0, \\
        \omega(\mathbf{x}, 0) &= \omega_0(\mathbf{x}).
    \end{aligned}
\end{equation}
Here, $\omega$ is the vorticity, $\mathbf{u}$ is the divergence-free velocity field, and $\nu=10^{-3}$ is the viscosity. The system is initialized with a random vorticity field $\omega_0$ drawn from a statistically stationary Gaussian distribution and is driven by a deterministic, large-scale forcing $f(\mathbf{x}) = 0.1(\sin(2\pi(x+y)) + \cos(2\pi(x+y)))$. A high-frequency stochastic term $\xi$, representing white-in-time noise with amplitude $\beta=5\times10^{-5}$, is included to mimic unresolved physical fluctuations.

Our data-driven closure task is to learn a model for the unresolved subgrid-scale dynamics. We define the closure term, $H$, as the combination of the nonlinear advection and the stochastic forcing, both of which are considered unknown to the coarse-grained model:
\begin{equation}\label{eq:convection}
    H(\mathbf{x},t) = -\mathbf{u}(\mathbf{x},t)\cdot\nabla\omega(\mathbf{x},t) + \beta\xi(\mathbf{x},t).
\end{equation}
The goal is to learn a generative model for the conditional distribution $p(H | \omega)$.

To generate the training dataset, we perform 100 independent high-fidelity simulations of Eq.~\eqref{eq:2dNS} on a fine $256 \times 256$ grid using a pseudo-spectral method with a Crank-Nicolson time-stepping scheme ($\Delta t = 10^{-3}$). To ensure the data represents a statistically stationary state, we discard the initial 30 seconds of each simulation. The remaining solutions are spatially downsampled to a coarse $64 \times 64$ grid and temporally subsampled at 0.01-second intervals. This process yields a final dataset of 20,000 pairs, consisting of the resolved vorticity fields $\omega$ (the conditional input) and the corresponding subgrid closure terms $H$ (the prediction target). This dataset is then split into 18,000 pairs for training and 2,000 for testing.

\subsection{Experimental Design and Comparative Framework}\label{ssec:exp_summary}
In the following sections, we present a systematic comparison to evaluate the performance of different transport-based generative closures. Our experimental framework is designed as a matrix of comparisons between three core generative paradigms and five distinct data representation and training strategies.

The three generative paradigms under investigation are:
\begin{enumerate}[label=(\roman*), topsep=3pt, itemsep=2pt]
    \item \textbf{Score-based Diffusion Models (DM):} An SDE-based stochastic approach that reverses a fixed noising process.
    \item \textbf{Flow Matching (FM):} A deterministic ODE-based approach that learns a velocity field along straight interpolation paths.
    \item \textbf{Stochastic Interpolants (SI):} A flexible hybrid framework. We evaluate this paradigm using two distinct source distributions: a standard Gaussian prior and an empirical prior derived from the conditioning variable itself.
\end{enumerate}

Each of these paradigms is applied across the following five data representation and training strategies:
\begin{enumerate}[label=(\roman*), topsep=3pt, itemsep=2pt]
    \item \textbf{Physical Space:} A baseline model operating directly on the full-resolution $64 \times 64$ fields.
    \item \textbf{Latent Space (No Regularization):} A two-phase model using a standard, reconstruction-only autoencoder on $16 \times 16$ latent fields.
    \item \textbf{Latent Space (Joint Training):} An end-to-end trained model with implicit regularization on the $16 \times 16$ latent fields.
    \item \textbf{Latent Space (Metric-Preserving):} A two-phase model with explicit MP regularization applied during autoencoder pre-training.
    \item \textbf{Latent Space (Geometry-Aware):} A two-phase model with explicit GA regularization applied during autoencoder pre-training.
\end{enumerate}
The performance of each combination is evaluated based on quantitative error metrics, the preservation of physical statistics (e.g., energy spectra), and computational cost.

\subsection{Performance in Physical Space: A Baseline for Comparison}\label{ssec:physical_space_compare}
This section establishes a performance baseline for the three transport-based generative paradigms. All models are trained and evaluated directly on the full-resolution $64 \times 64$ physical-space data. We denote these models with a "P-" prefix (e.g., P-DM, P-FM, P-SI) to distinguish them from the latent-space variants analyzed in subsequent sections.

To assess both the predictive accuracy and the uncertainty representation of each paradigm, we perform an ensemble-based analysis. For a given conditional input $\omega$, we draw an ensemble of $N_s = 1000$ closure samples $\{\tilde{H}_i\}_{i=1}^{N_s}$ from the learned conditional distribution $p_\theta(H|\omega)$. The accuracy of the model is evaluated using the ensemble mean, $\bar{H} = \frac{1}{N_s} \sum_{i=1}^{N_s} \tilde{H}_i$, which represents the model's deterministic best guess. The model's predicted uncertainty is characterized by the pointwise standard deviation of the ensemble, which is compared against the ground-truth variability.

We quantify the error of the ensemble-mean prediction against the ground truth $H$ using two metrics: the Mean Squared Error (MSE) and the Relative Error (RE), defined as:
\begin{equation}
    D_{\mathrm{MSE}} = \frac{1}{N_p} \| \bar{H} - H \|_F^2,
\end{equation}
and
\begin{equation}
    D_{\mathrm{RE}} = \frac{\| \bar{H} - H \|_F}{\| H \|_F},
\end{equation}
where $\| \cdot \|_F$ is the Frobenius norm and $N_p$ is the total number of grid points in the field.

\begin{figure}[!htb]
    \centering
    \includegraphics[width=\linewidth]{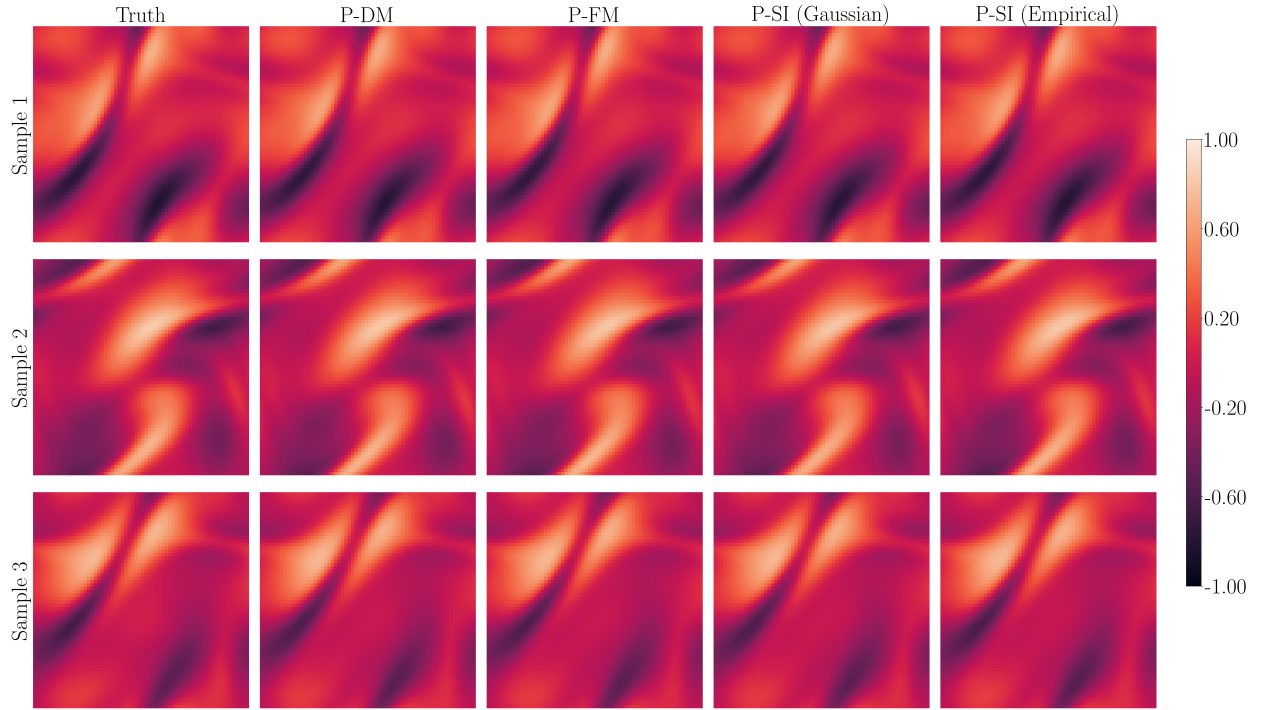}
    \caption{
    Qualitative comparison of stochastic closure samples from physical-space models. 
    This figure assesses the performance of conditional generation. Each column corresponds to a different model: the ground truth, P-DM, P-FM, and P-SI with two different priors. Each row displays an independent, random sample of the closure term $H$, all generated for the same input vorticity field $\omega$.}
    \label{fig:physical_samples}
\end{figure}

We first evaluate the qualitative performance of the physical-space models. As shown in Figure~\ref{fig:physical_samples}, all three paradigms (DM, FM, and SI) generate high-fidelity, diverse samples of the closure term $H$. The generated samples are structurally consistent with the ground truth, demonstrating that the models have learned a meaningful conditional distribution. 

\begin{table}[!htb]
\centering
\caption{
    Quantitative comparison of physical-space generative models.
    All metrics are averaged over the test set. Per-sample errors measure the average error of individual stochastic draws, while ensemble-mean errors measure the accuracy of the averaged prediction. Field Std. is the spatially-averaged standard deviation of the generated ensemble, indicating the magnitude of modeled uncertainty. Values in gray denote $\pm$ two standard deviation over the test set instances.
}
\begin{adjustbox}{width=\linewidth}
\begin{tabular}{@{}cccccc@{}}
\toprule
\multirow{2}{*}{\textbf{Models}} & \multicolumn{2}{c}{\textbf{Mean of per-sample errors}}         & \multicolumn{2}{c}{\textbf{Error of ensemble mean}}      & \multirow{2}{*}{\textbf{Field Std.}} \\ 
\cmidrule(lr){2-5}
& $D_\text{MSE}^{\text{sample}}$ & $D_\text{RE}^{\text{sample}}$ & $D_\text{MSE}^{\text{ens}}$ & $D_\text{RE}^{\text{ens}}$ &                                      \\ 
\midrule
P-DM                 & 8.231e-04 {\footnotesize\color{gray} $\pm$ 1.185e-04} & 1.157e-01 {\footnotesize\color{gray} $\pm$ 8.451e-03} & 4.696e-04 & 8.725e-02 & 1.885e-02 {\footnotesize\color{gray} $\pm$ 2.702e-03} \\

P-FM                  & 9.340e-04 {\footnotesize\color{gray} $\pm$ 1.165e-04} & 1.231e-01 {\footnotesize\color{gray} $\pm$ 7.484e-03} & 5.051e-04 & 9.028e-02 & 2.074e-02 {\footnotesize\color{gray} $\pm$ 3.390e-03} \\

P-SI (Gaussian)              & 8.831e-04 {\footnotesize\color{gray} $\pm$ 1.184e-04} & 1.199e-01 {\footnotesize\color{gray} $\pm$ 7.951e-03} & 4.541e-04 & 8.595e-02 & 2.076e-02 {\footnotesize\color{gray} $\pm$ 3.162e-03} \\
                      
P-SI (Empirical)             & 8.732e-04 {\footnotesize\color{gray} $\pm$ 1.266e-04} & 1.192e-01 {\footnotesize\color{gray} $\pm$ 8.591e-03} & 4.566e-04 & 8.610e-02 & 2.050e-02 {\footnotesize\color{gray} $\pm$ 2.501e-03} \\

\bottomrule
\end{tabular}
\end{adjustbox}
\label{Tab: comparison_ensemble_physics}
\end{table}

For a quantitative analysis, Table~\ref{Tab: comparison_ensemble_physics} reports several key metrics. The ensemble-mean errors ($D_{\text{RE}}^{\text{ens}}$), which measure how well each model captures the deterministic component of the closure, are comparable across all methods, ranging from 8.6\% to 9.0\%. This confirms that all paradigms are highly effective at this task, with the P-SI models showing a marginal advantage. As expected, these errors are consistently lower than the average per-sample errors due to the variance-reduction effect of averaging.

Crucially, the table also allows us to evaluate how well the models represent the prescribed stochasticity via the spatially-averaged standard deviation (Field Std.). For our problem setup, this quantity has an analytical reference value of 0.02 (see~\ref{ssec:q_wiener}). All models reproduce this target with high fidelity; the P-FM and P-SI models match the ground-truth variance almost exactly, while the P-DM slightly underestimates it, though all are within a 10\% relative error. Accurately capturing the uncertainty level is critical for preserving the system's physical statistics in forward simulations. These combined results establish that all physical-space generative paradigms perform well, accurately capturing both the mean behavior and the uncertainty level of the closure.

\subsection{The Role of Transport Geometry in Sampling Efficiency}\label{ssec:physical_space_straightness}
While the previous section established that all generative paradigms achieve comparable accuracy with a sufficient number of sampling steps, their computational efficiency varies dramatically. As shown in Table~\ref{tab: mse_re_comparison_efficiency}, reducing the number of integration steps causes the accuracy of the P-DM model to degrade sharply, with a catastrophic failure at a single step. In contrast, the linear-interpolation-based P-FM and P-SI models remain remarkably stable, with P-FM showing almost no loss in accuracy even in the single-step regime.

\begin{table}[!htb]
\caption{
    Sampling accuracy vs. number of integration steps. Linear-interpolation–based methods (P-FM, P-SI) remain stable with far fewer steps, while the P-DM model degrades sharply under coarse discretization.
}
\centering
\resizebox{\textwidth}{!}{%
\begin{tabular}{@{}lcccccccc@{}}
\toprule
\multirow{3}{*}{\textbf{Sample steps}}
& \multicolumn{2}{c}{\multirow{2}{*}{\textbf{P-DM}}} 
& \multicolumn{2}{c}{\multirow{2}{*}{\textbf{P-FM}}} 
& \multicolumn{4}{c}{\textbf{P-SI}} \\

& &
& & 
& \multicolumn{2}{c}{Gaussian priors} 
& \multicolumn{2}{c}{Empirical priors} \\
\cmidrule(lr){2-3} \cmidrule(lr){4-5} \cmidrule(lr){6-7} \cmidrule(lr){8-9}
&$D_\mathrm{MSE}$ & $D_\mathrm{RE}$ &$D_\mathrm{MSE}$ & $D_\mathrm{RE}$ & $D_\mathrm{MSE}$ & $D_\mathrm{RE}$ & $D_\mathrm{MSE}$ & $D_\mathrm{RE}$\\
\midrule
100  & 7.945e-04 & 1.197e-01 & 7.228e-04 & 1.112e-01 & 8.776e-04 & 1.254e-01 & 8.576e-04 & 1.244e-01  \\
50   & 7.431e-03 & 3.158e-01 & 8.025e-04 & 1.197e-01 & 8.736e-04 & 1.250e-01 & 8.641e-04 & 1.249e-01  \\
10   & 3.130e-02 & 7.563e-01 & 8.796e-04 & 1.200e-01 & 8.971e-04 & 1.269e-01 & 9.390e-04 & 1.298e-01  \\
1    & 8.262e+02 & 2.851e+02 & 9.846e-04 & 1.334e-01 & 3.434e-03 & 2.502e-01 & 2.845e-03 & 2.265e-01  \\
\bottomrule
\end{tabular}%
}
\label{tab: mse_re_comparison_efficiency}
\end{table}

This difference in robustness is a direct consequence of the underlying transport path geometry. To quantify this, we define a straightness metric $\mathcal{S} \in [0,1]$ that measures the ratio of the direct Euclidean distance between a trajectory's start and end points to the total integrated path length (a value of $\mathcal{S}=1$ indicates a perfectly straight line). Table~\ref{tab:straightness_comparison} confirms the link between path geometry and sampler stability. P-FM follows almost perfectly linear paths ($\mathcal{S} \approx 0.999$), explaining its tolerance to large step sizes. In contrast, P-DM traces highly curved paths ($\mathcal{S} < 0.3$), which require fine discretization to integrate accurately. P-SI occupies a middle ground, consistent with its moderate stability. The $\mathcal{S}=1$ values for all models at a single step are a geometric artifact, as a one-step path is trivially straight but not necessarily accurate. Thus, high $\mathcal{S}$ must be interpreted in conjunction with accuracy metrics to assess practical efficiency.


\begin{table}[!htb]
\caption{
    Trajectory straightness ($\mathcal{S}$) for different sampling methods. Higher values ($\to 1$) indicate straighter paths. P-FM's near-perfect straightness explains its robustness to coarse time discretization.
}
\centering
\resizebox{\textwidth}{!}{%
\begin{tabular}{@{}lcccccccc@{}}
\toprule
\multirow{2}{*}{\textbf{Sample steps}}
& \multicolumn{2}{c}{\multirow{2}{*}{\textbf{P-DM}}} 
& \multicolumn{2}{c}{\multirow{2}{*}{\textbf{P-FM}}}
& \multicolumn{4}{c}{\textbf{P-SI}} \\
& &
& & 
& \multicolumn{2}{c}{Gaussian priors} 
& \multicolumn{2}{c}{Empirical priors} \\
\midrule

100  & 1.278e-01 & {\footnotesize\color{gray} $\pm$ 2.467e-03}
     & 9.995e-01 & {\footnotesize\color{gray} $\pm$ 1.881e-04}
     & 1.929e-01 & {\footnotesize\color{gray} $\pm$ 3.938e-03}
     & 1.857e-01 & {\footnotesize\color{gray} $\pm$ 1.357e-02} \\
50   & 1.717e-01 & {\footnotesize\color{gray} $\pm$ 3.293e-03}
     & 9.995e-01 & {\footnotesize\color{gray} $\pm$ 1.718e-04}
     & 2.602e-01 & {\footnotesize\color{gray} $\pm$ 5.120e-03}
     & 2.508e-01 & {\footnotesize\color{gray} $\pm$ 1.770e-02} \\
10   & 2.553e-01 & {\footnotesize\color{gray} $\pm$ 4.854e-03}
     & 9.997e-01 & {\footnotesize\color{gray} $\pm$ 1.153e-04} 
     & 4.524e-01 & {\footnotesize\color{gray} $\pm$ 8.342e-03}
     & 4.373e-01 & {\footnotesize\color{gray} $\pm$ 2.716e-02}
     \\
1    & 1.000e+00 & {\footnotesize\color{gray} $\pm$ 0.000e+00}
     & 1.000e+00 & {\footnotesize\color{gray} $\pm$ 0.000e+00}
     & 1.000e+00 & {\footnotesize\color{gray} $\pm$ 0.000e+00}
     & 1.000e+00 & {\footnotesize\color{gray} $\pm$ 0.000e+00} \\
\bottomrule
\end{tabular}%
}
\label{tab:straightness_comparison}
\end{table}

\begin{figure}[!htb]
    \centering
    \includegraphics[width=\linewidth]{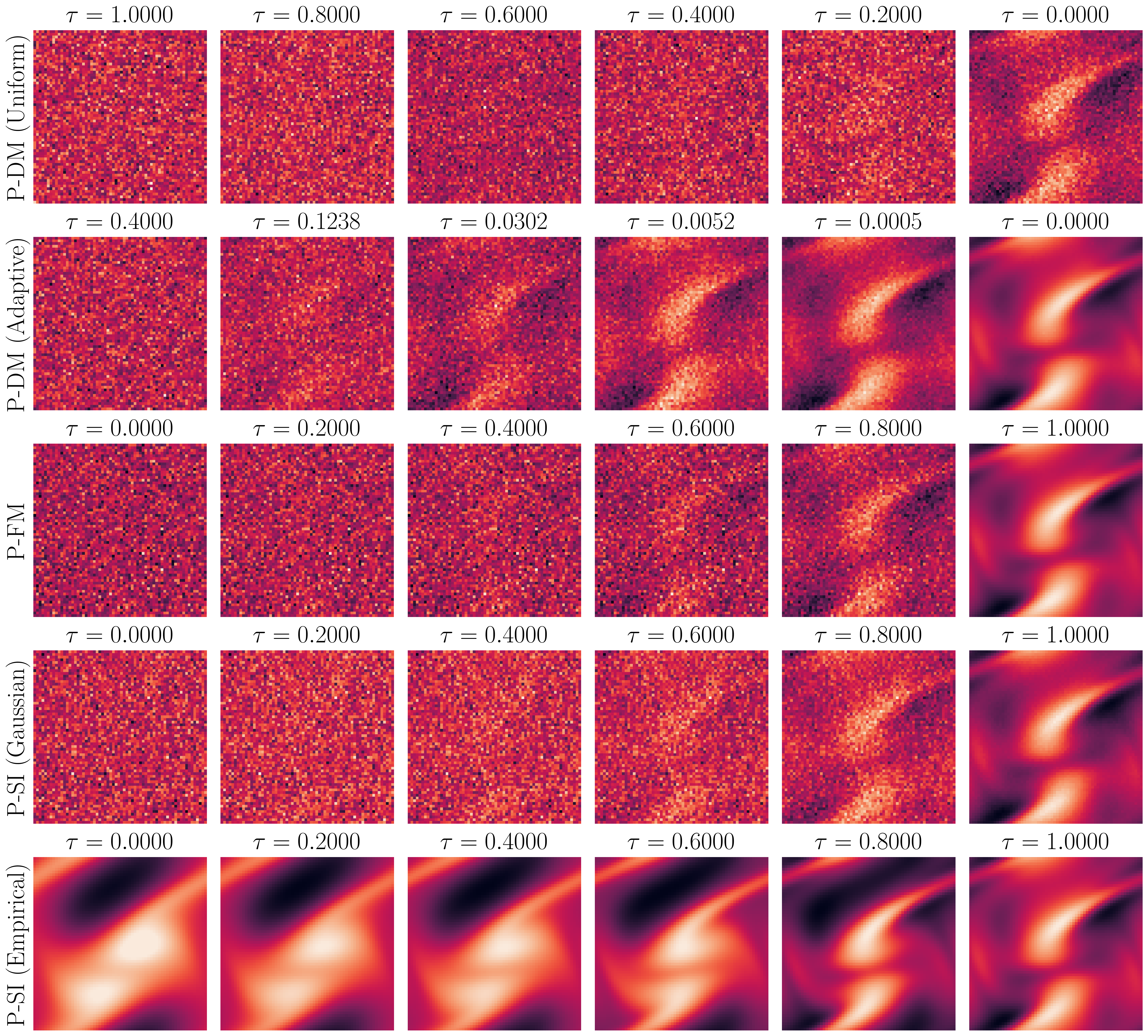}
    \caption{
        Visualization of 10-step sampling trajectories.
        Each row shows intermediate states of the generated field for a different sampling strategy, with time $\tau$ evolving according to the model's process. The comparison between the standard P-DM (Row 1) and an adaptive P-DM with reduced initial variance (Row 2) highlights how sampler design can overcome path curvature. The smooth evolution of P-FM (Row 3) and P-SI (Rows 4-5) visually confirms their straighter transport paths.
    }
    \label{fig:closure_field_sampling}
\end{figure}

Figure~\ref{fig:closure_field_sampling} visualizes these distinct dynamics and provides crucial insights into sampler design. The standard P-DM sampler with uniform steps (Row 1) fails to converge in a 10-step budget, as its highly curved path cannot be integrated accurately with coarse steps. However, the performance of diffusion models can be dramatically improved with a more tailored strategy. Row 2 shows an adaptive P-DM sampler that successfully converges in the same 10-step budget. This is achieved by:
\begin{enumerate}
    \item Reducing Initial Variance: Since we are targeting a narrow conditional distribution, the reverse SDE can be initialized from a state of lower noise (i.e., starting at $\tau < \Tau$), significantly shortening the required path length.
    \item Adaptive Time-Stepping: An adaptive schedule uses larger steps in the high-noise regime where the path is smoother and smaller steps near the data manifold ($\tau \to 0$) where curvature and stiffness increase, thus optimizing the discretization.
\end{enumerate}
In contrast, the inherently straighter paths of P-FM (Row 3) and P-SI (Rows 4-5) allow them to converge smoothly even with simple uniform time-stepping, reinforcing the conclusion that path geometry is a primary determinant of sampling efficiency.

\subsection{Impact of Latent Space Geometry on Generative Performance}\label{ssec:latent_space_comparisons}
Latent generative models offer significant computational advantages, but their success is critically dependent on the quality of the latent space. In this section, we systematically evaluate how different autoencoder (AE) training strategies impact the geometry of the latent space and the performance of the downstream generative closure model. We compare four families of strategies: two-phase training without regularization (NoReg), end-to-end joint training (Joint), and two-phase training with explicit metric-preserving (MP) or geometry-aware (GA) regularization.

\subsubsection{The Failure of Reconstruction-Only Latent Spaces}
We first establish that a latent space optimized solely for reconstruction is unsuitable for generative modeling. As shown in Table~\ref{tab: generation_compare_latent}, the NoReg strategy results in catastrophic failure, with a latent-space relative error ($D_{\text{RE}}$) of approximately 30\%, which translates to a decoded physical-space error of 33\%—far worse than any of the physical-space baseline models.

\begin{figure}[!htb]
    \centering
    \includegraphics[width=\linewidth]{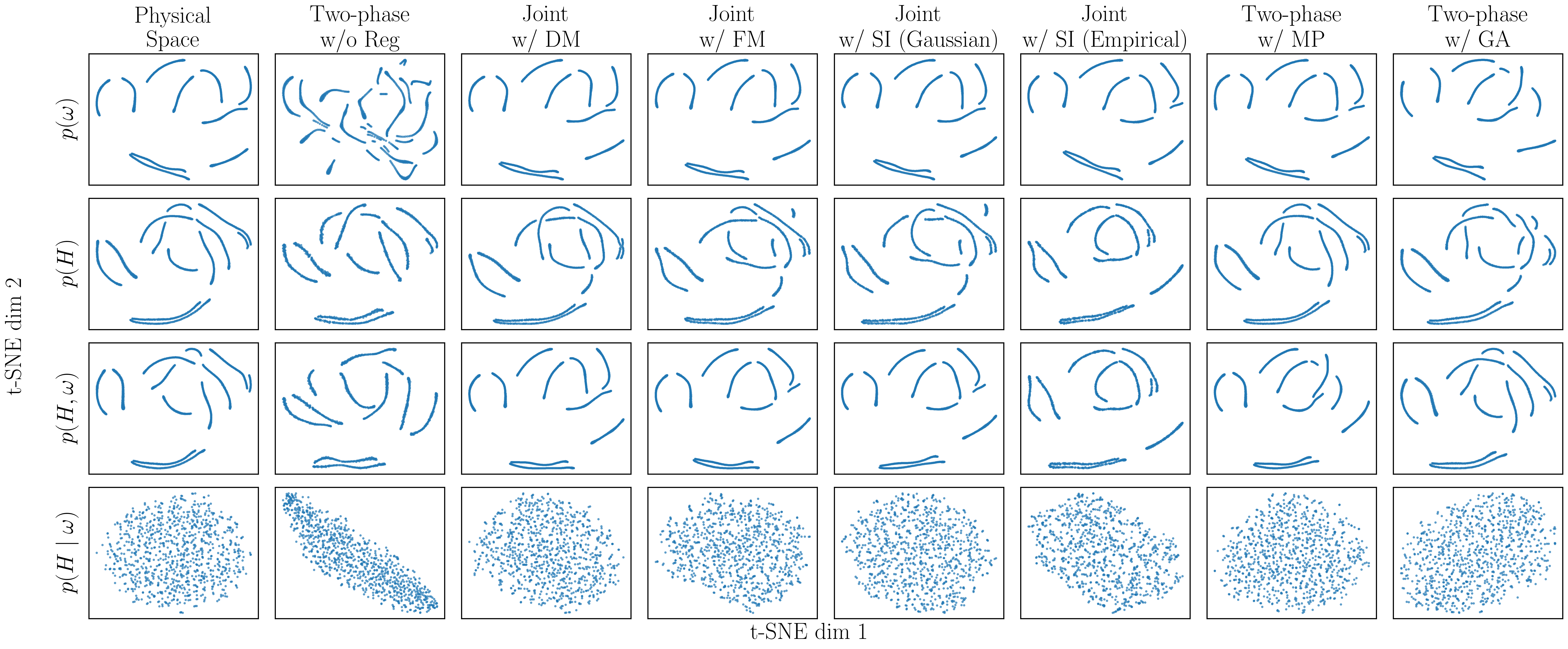}
        \caption{
        t-SNE visualizations of latent space structure. Each column represents a different training strategy. The unregularized latent space (Column 2) is visibly distorted compared to the physical space ground truth (Column 1). Joint training (Columns 3-6) and explicit regularization (Columns 7-8) produce far more coherent structures.
    }
    \label{fig: tsne_latent}
\end{figure}

The reason for this failure is revealed by the latent space geometry, visualized in Figure~\ref{fig: tsne_latent} and quantified in Table~\ref{tab: distortion_metrics}. The NoReg autoencoder produces a highly distorted latent space, evidenced by its scattered t-SNE embeddings, large Procrustes Disparity (PD), and a nearly threefold increase in the conditional coefficient of variation (CV) compared to the physical space. Crucially, this geometric failure occurs despite the NoReg model achieving the lowest reconstruction error of all tested strategies (Table~\ref{tab:ae_recon_omega_H}). This establishes our central thesis: faithful reconstruction is a necessary but insufficient condition for effective latent generative modeling; the geometry of the latent space is paramount.

\subsubsection{The Efficacy of Implicit and Explicit Regularization}
Structuring the latent space, either implicitly or explicitly, dramatically improves performance. Joint training provides a powerful implicit regularization, forcing the AE to co-adapt with the generative model. This alignment yields a massive improvement over the NoReg baseline, reducing the latent-space error by over 4x and achieving a final physical-space accuracy comparable to the physical-space models (Table~\ref{tab: generation_compare_latent}). This performance gain is directly linked to an improved latent geometry with lower distortion metrics shown in Table~\ref{tab: distortion_metrics}.

However, explicit geometric regularization during a two-phase pipeline proves to be the most effective and stable strategy. The results in Table~\ref{tab: generation_compare_latent} are unequivocal: the MP and GA regularized models achieve the lowest latent-space generation errors of all methods. The MP-regularized models are the top performers, yielding the lowest latent-space error ($\sim$2.9\%) and a final physical-space error ($\sim$8.3\%) that matches or even slightly surpasses the physical-space baselines.

\begin{figure}[!htb]
    \centering
    \includegraphics[width=0.7\linewidth]{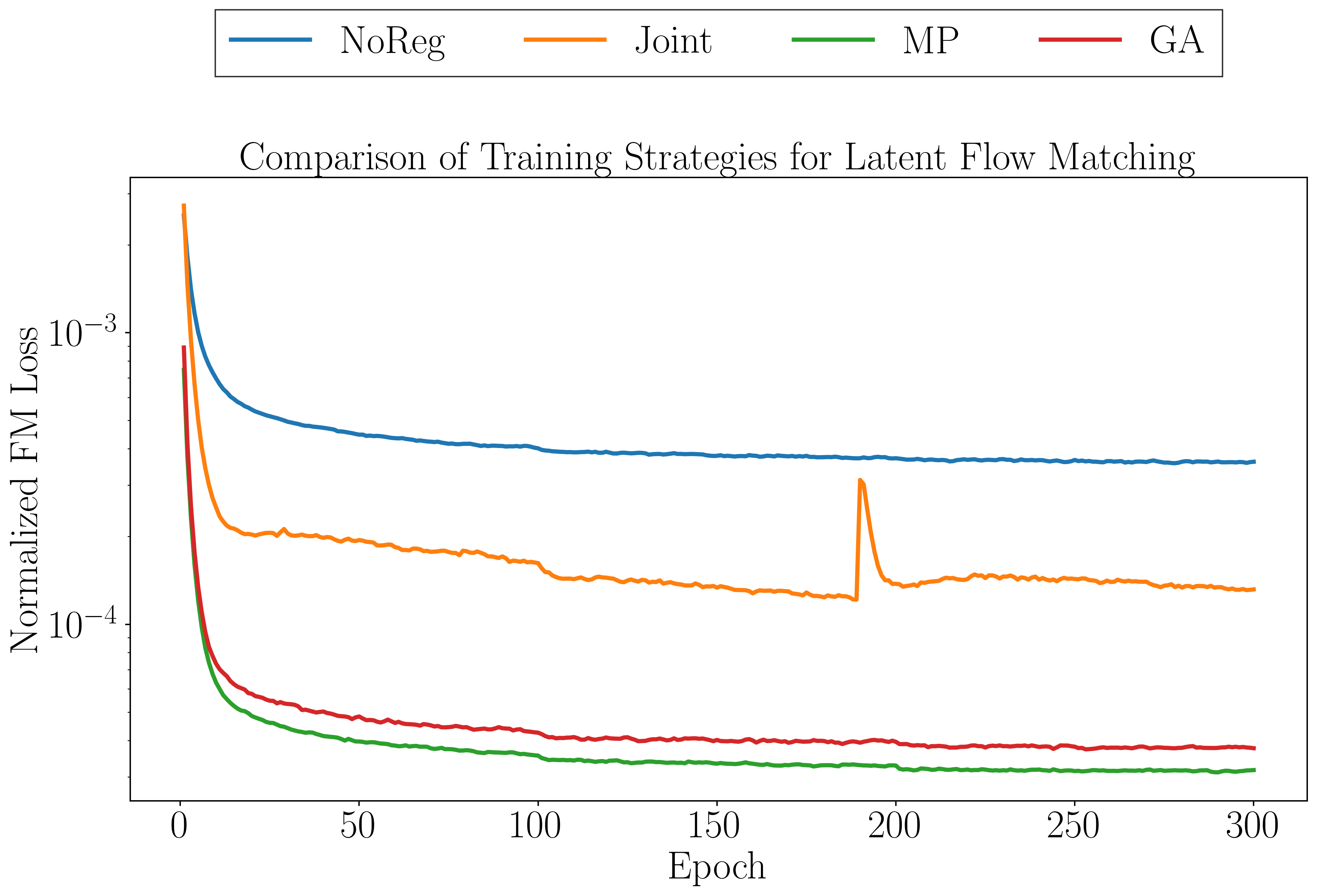}
    \caption{
        Normalized Flow Matching (FM) training loss (log-scale).
        The explicitly regularized models (MP, GA) achieve the lowest and smoothest loss trajectories, indicating that a well-structured latent space simplifies the generative learning task. The Joint model exhibits a non-stationary spike, while the NoReg model converges to a much higher loss.
    }
    \label{fig:training_loss}
\end{figure}

The superiority of the explicitly regularized models is corroborated by the training loss curves in Figure~\ref{fig:training_loss}, which use a normalized Flow Matching (FM) loss for direct comparison. The smooth, low-loss trajectories of the MP and GA models indicate that a well-structured latent space makes the generative objective fundamentally easier to optimize. In contrast, the Joint model's loss curve, while better than NoReg, exhibits a pronounced spike attributed to the non-stationarity of the co-adapting AE.

Of the two explicit regularizers, MP consistently outperforms GA in generative tasks. The advantage of MP arises from its direct alignment with the mechanics of the transport-based samplers. Operations such as discretized SDE/ODE integration, interpolation, and gradient evaluation are performed in Euclidean coordinates; preserving local Euclidean distances (MP) is therefore more beneficial than preserving geodesic distances (GA). Furthermore, MP is more computationally efficient, as it requires only pairwise Euclidean distances rather than more expensive geodesic estimates. In summary, for latent generative closures, explicitly regularizing the AE geometry with metric-preserving constraints is the most effective, stable, and efficient strategy.



\begin{table}[!htb]
\centering
\caption{
    Autoencoder reconstruction errors.
    Each block reports mean-squared ($D_{\mathrm{MSE}}$) and relative ($D_{\mathrm{RE}}$) errors. Note that the Recon only baseline has the lowest reconstruction error, yet the worst generative performance (see Table~\ref{tab: generation_compare_latent}).
}
\begin{adjustbox}{width=0.8\linewidth}
\begin{tabular}{@{}l c cc cc@{}}
\toprule
\multicolumn{2}{c}{\multirow{2}{*}{\textbf{Autoencoders}}}      & 
\multicolumn{2}{c}{\textbf{Vorticity} $\omega$} & 
\multicolumn{2}{c}{\textbf{Convection} $H$} \\
\cmidrule(lr){3-4} \cmidrule(lr){5-6}
&  & $D_{\mathrm{MSE}}$ & $D_{\mathrm{RE}}$ & $D_{\mathrm{MSE}}$ & $D_{\mathrm{RE}}$ \\
\midrule
Recon only & -         & 5.280e-07 & 7.382e-04 & 7.168e-06 & 1.124e-02 \\ 
\cmidrule(lr){2-6}
\multirow{4}{*}{Joint-trained}                       
& w/ DM           & 5.772e-06 & 2.484e-03 & 4.019e-05 & 2.439e-02 \\
& w/ FM & 4.638e-06 &2.215e-03 & 6.265e-06 & 1.039e-02 \\
& w/ SI (Gaussian) & 5.041e-05 & 7.231e-03 & 3.180e-05 & 2.219e-02 \\
& w/ SI (Empirical)& 5.133e-06 & 2.334e-03 & 4.419e-06 & 8.650e-03 \\
\cmidrule(lr){2-6}
\multirow{2}{*}{Regularized}                 
& MP  & 5.571e-05 & 7.666e-03 & 2.415e-05 & 1.844e-02 \\
& GA  & 3.920e-05 & 6.461e-03 & 2.592e-05 & 1.921e-02 \\
\bottomrule
\end{tabular}
\end{adjustbox}
\label{tab:ae_recon_omega_H}
\end{table}

\begin{table}[!htb]
\caption{
    Generative performance across different latent space strategies.
    Physical-space models (P-) are baselines. Latent-space models (L-) show that NoReg fails, Joint improves significantly, and explicit regularization (MP, GA) performs best. Bold values highlight the best performance within each category (latent and physical errors for MP).
}
\centering
\begin{adjustbox}{width=0.9\linewidth}
\begin{tabular}{@{}ccccccc@{}}
\toprule
\multicolumn{3}{c}{\multirow{2}{*}{\textbf{Model}}}                                                                                              & \multicolumn{2}{c}{\textbf{Latent space}} & \multicolumn{2}{c}{\textbf{Physical space}}                                               \\
\multicolumn{3}{c}{}                                                                                                                    & \multicolumn{2}{c}{\textbf{generation}}   & \multicolumn{2}{c}{\textbf{generation}}                                                                             \\ \cmidrule(lr){1-3}
\cmidrule(lr){4-5} \cmidrule(lr){6-7}
\multicolumn{2}{c}{Catagory}                                                                                          & Paradigms       & $D_\text{MSE}$       & $D_\text{RE}$      & $D_\text{MSE}$        & $D_\text{RE}$                                                                   \\ \midrule
\multicolumn{2}{l}{\multirow{4}{*}{Physical space models}}                                                            & P-DM           & -                    & -                  & 4.696e-04             & 8.725e-02            \\

\multicolumn{2}{c}{}                                                                                                  & P-FM & -                    & -                  & 4.911e-04             & 8.904e-02            \\ 

\multicolumn{2}{c}{}                                                                                                  & P-SI (Gaussian) & -                    & -                  & 4.612e-04             & 8.662e-02            \\ 

\multicolumn{2}{c}{}                                                                                                  & P-SI (Empirical) & -                    & -                  & 4.994e-04             & 9.032e-02            \\ 

\cmidrule(l){3-7} 
\multirow{16}{*}{Latent space models} & \multirow{4}{*}{\begin{tabular}[c]{@{}c@{}}Two-phase \\ w/o regs\end{tabular}} & L-DM           & 9.524e-02            & 3.154e-01          & 6.220e-03             & 3.290e-01            \\

                                     &                                                                                & L-FM & 9.501e-02            & 3.045e-01          & 6.679e-03             & 3.302e-01            \\ 

                                     &                                                                                & L-SI (Gaussian) & 1.098e-01            & 3.171e-01          & 7.030e-03             & 3.379e-01            \\ 
                                        &               & L-SI (Empirical) & 9.637e-02            & 2.995e-01          & 6.716e-03             & 3.313e-01            \\ 
                                     
                                    \cmidrule(l){3-7} 
                                     & \multirow{4}{*}{Joint-trained}                                                 
                                     & L-DM           & 3.677e-05            & 7.630e-02          & 4.877e-04             & 8.932e-02            \\
                                                                          &                                                                                
                                     & L-FM & 3.525e-03            & 7.394e-02         & 4.862e-04             & 8.928e-02            \\ 
                                     
                                     &                                                                                
                                     & L-SI (Gaussian) & 2.812e-03            & 7.290e-02          & 5.131e-04             & 9.151e-02            \\ 
                                     &               
                                     & L-SI (Empirical) & 3.129e-03           & 8.274e-01          & 6.715e-04             & 9.219e-01            \\

                                                                          \cmidrule(l){3-7} 
                                     & \multirow{4}{*}{\begin{tabular}[c]{@{}c@{}}Two-phase \\ w/ MP\end{tabular}}                                                 
                                     & L-DM           & 6.379e-03            & \textbf{2.936e-02}          & 4.224e-04             & \textbf{8.279e-02}            \\
                                                                          &     
                                     & L-FM & 6.365e-03            & \textbf{2.932e-02}          & 4.243e-04             & \textbf{8.299e-02}            \\ 
                                     
                                     &                                                                                
                                     & L-SI (Gaussian) & 6.592e-03            & \textbf{2.982e-02}          & 4.400e-04            & \textbf{8.456e-02}            \\ 
                                     &               
                                     & L-SI (Empirical) & 6.297e-03            & \textbf{2.852e-02}          & 4.302e-04            & \textbf{8.322e-02}           \\ 

                                                                          \cmidrule(l){3-7} 
                                     & \multirow{4}{*}{\begin{tabular}[c]{@{}c@{}}Two-phase \\ w/ GA\end{tabular}}                                                    & L-CDM           & 2.275e-02            & 3.801e-02          & 6.006e-04            & 9.986e-02            \\
                                                                         &  
                                      & L-FM & 1.851e-02            & 3.439e-02          & 5.642e-04             & 9.673e-02            \\ 
                                     
                                     &        
                                                                           
                                     & L-SI (Gaussian)  & 1.869e-02             & 3.457e-02          & 5.614e-04             & 9.660e-02            \\ 
                                     &               
                                     & L-SI (Empirical) & 2.223e-02            & 3.772e-02         & 5.560e-04             & 9.526e-02             \\ 
                                     
                                     \bottomrule
\end{tabular}
\end{adjustbox}
\label{tab: generation_compare_latent}
\end{table}

\begin{table}[!htb]
\centering
\caption{Quantitative assessment of latent space distortion using Procrustes disparity (PD) and mean of absolute CV. PD measures geometric dissimilarity between t-SNE embeddings after optimal alignment—lower values indicate better preservation of data structure. Mean of absolute CV quantifies the relative spread of conditional distributions $p(H\mid \omega)$—lower values indicate more compact conditionals. Bold values indicate best performance per metric.}
\begin{adjustbox}{width=\linewidth}
\begin{tabular}{@{}l c c c c c c c c c@{}}
\toprule
\multirow{3}{*}{\textbf{Distributions}} 
  & \multirow{3}{*}{\textbf{Metrics}} 
  & \multicolumn{8}{c}{\textbf{Spaces}}                                                     \\ 
\cmidrule(l){3-10}
  & & Physical & Two-phase        & Joint    & Joint    & Joint  & Joint      & Two-phase    & Two-phase    \\
  & & Space    & w/o Reg       & w/ DM  & w/ FM  & w/ SI (Gaussian)    & w/ SI (Empirical)   & w/ MP     & w/ GA     \\ 
\midrule
$p(\omega)$   
  & PD 
  & –          & 5.370e-01 
  & 9.855e-03 & \textbf{1.215e-02}   & 1.434e-02 
  & 1.371e-02                   & 1.561e-02  & 6.710e-02       \\ 
  
$p(H)$        
  & PD 
  & –          & 1.637e-01 
  & 2.130e-02 & 2.835e-02   & 4.302e-02 
  & 9.609e-02                   & \textbf{3.772e-03}  & 3.938e-02       \\ 
$p(H,\omega)$
  & PD 
  & –          & 2.567e-01 
  & 1.003e-01 & 8.162e-02   & 7.826e-02 
  & 1.016e-01                   & 1.024e-01  & \textbf{9.534e-03}       \\ 
\midrule
$p(H\mid \omega)$
  & Mean of absolute CV 
  & 2.571e-01  & 7.562e-01 
  & 1.245e-01   & 2.378e-01
  & 2.365e-01 
  & 1.9092e-01  & \textbf{5.147e-02}  & 1.074e-01       \\ 
\bottomrule
\end{tabular}
\end{adjustbox}
\label{tab: distortion_metrics}
\end{table}

\subsection{A Posteriori Validation via Numerical Simulation}\label{ssec: closure_sim}

To assess the \textit{a posteriori} performance of the generative closure framework, we embed the trained models within a numerical solver for the 2-D Navier-Stokes equations (Eq.~\eqref{eq:2dNS}). The simulations are initialized at $t_0=30$ using a ground-truth vorticity field, $\omega(\mathbf{x}, t_0)$, and are integrated forward to $t=50$ with a time step of $\Delta t=10^{-3}$. The solver employs the same pseudo-spectral and Crank-Nicolson methods used to generate the training data. The system is evolved with a viscosity of $\nu = 10^{-3}$ and is subject to the deterministic forcing term $f(\mathbf{x}) = 0.1 (\sin(2\pi(x+y)) + \cos(2\pi(x+y)))$. For computational efficiency, the closure term is updated every five physical time steps.

We evaluate two strategies for deploying the stochastic closure:

\textbf{(1) Stochastic Trajectory Simulation.} 
To characterize the uncertainty propagation of the closure, we perform a Monte Carlo analysis. For each of $N_e=1000$ independent simulations, a single closure term $\tilde{H}(\mathbf{x}, t)$ is sampled from the conditional distribution $p(H\mid \omega)$ at each evaluation step. The resulting ensemble of trajectories allows for the computation of statistics, such as the mean and standard deviation of prediction errors.

\textbf{(2) Conditional Mean Simulation.} 
To obtain a deterministic prediction that minimizes variance, we approximate the conditional expectation of the closure term. At each evaluation step, we draw a large ensemble of $N_e=1000$ samples, $\{\tilde{H}_i\}_{i=1}^{N_e} \sim p(H\mid \omega)$, and compute their mean, $\bar{H} = \frac{1}{N_e} \sum_{i=1}^{N_e} \tilde{H}_i$. This mean value is then used as the closure term for a single simulation trajectory.

\begin{figure}[!htb]
    \centering
    \includegraphics[width=0.6\linewidth]{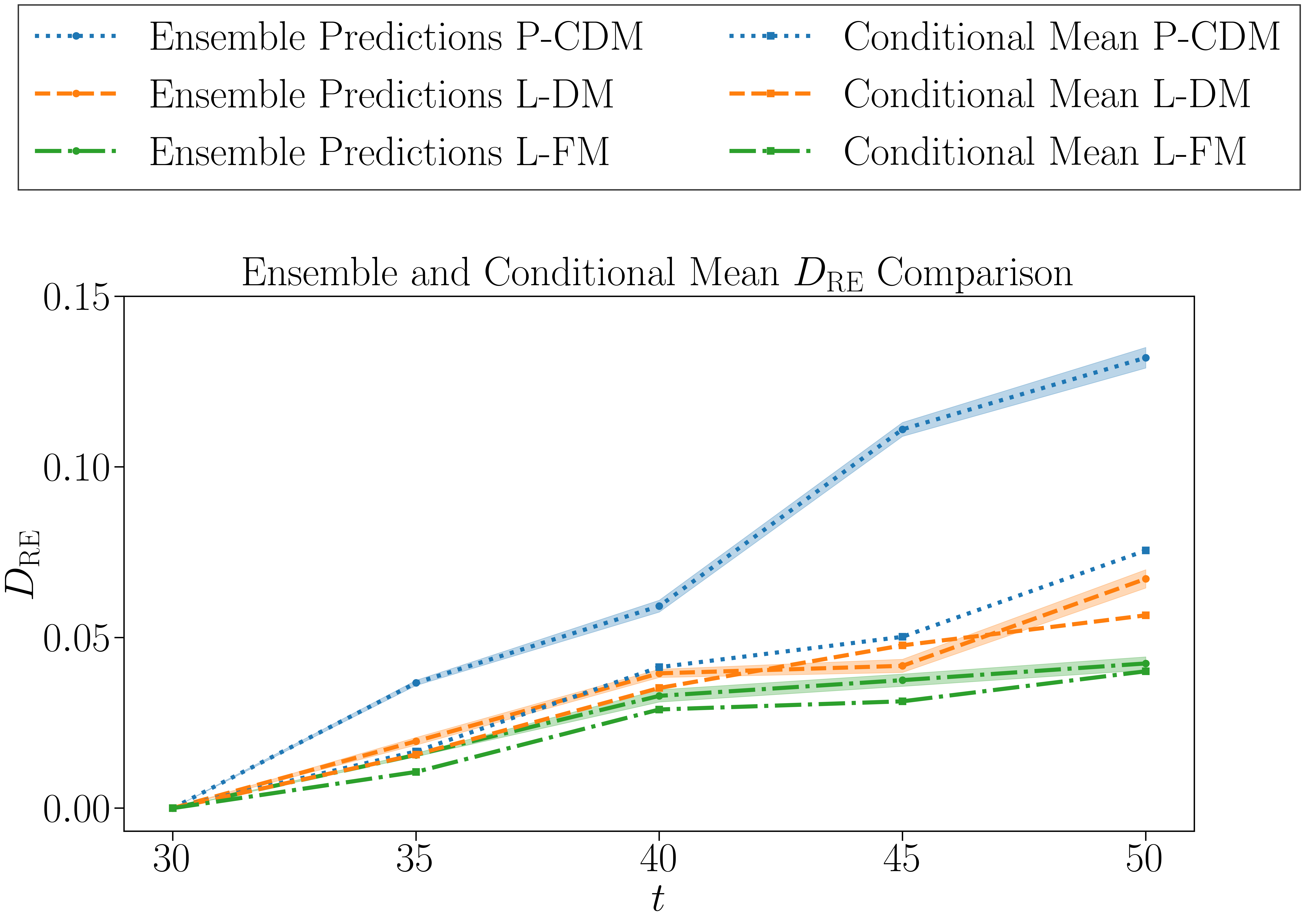}
    \caption{Temporal evolution of relative simulation error ($D_{\mathrm{RE}}$). Comparison of stochastic trajectories (dotted lines, representing the mean over 1000 runs) and ensemble-mean predictions (dashed/solid lines) for various closure models. All generative closures significantly outperform the uncorrected baseline (not shown, error reaches 0.84), with the L-FM model achieving the lowest error.}
    \label{fig: mse_re}
\end{figure}

\begin{figure}[!htb]
    \centering
    \includegraphics[width=\linewidth]{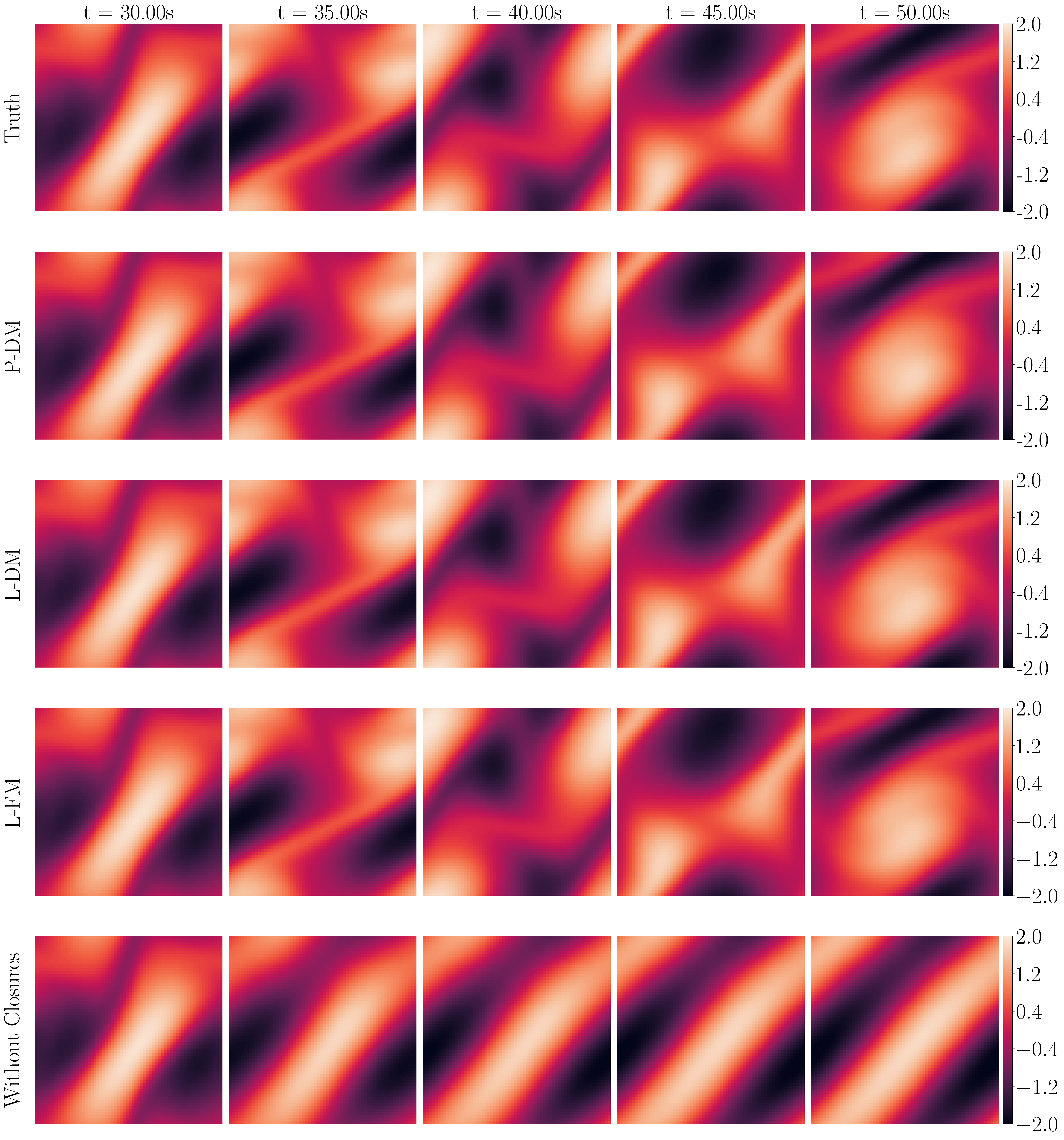}
    \caption{
        Qualitative comparison of vorticity field evolution. Snapshots from $t=30$ to $t=50$ for the high-fidelity ground truth, three generative closure models, and the uncorrected simulation. The closure-corrected simulations successfully capture the fine-scale flow structures, whereas the uncorrected simulation diverges and develops spurious features.
    }
    \label{fig: closure_h}
\end{figure}

Results summarized in Table~\ref{tab: simulation_metrics} and Figure~\ref{fig: mse_re} confirm the critical role of the closure term. The uncorrected simulation exhibits rapid error accumulation, with the relative error ($D_{\mathrm{RE}}$) reaching 84.2\% by $t=50$. In contrast, all generative closure models successfully mitigate this error growth, maintaining a final error below 13\%. The qualitative impact of this correction is visualized in Figure~\ref{fig: closure_h}, where the vorticity field from the corrected simulation remains structurally coherent and aligned with the high-fidelity reference, unlike the uncorrected field which diverges significantly in both pattern and magnitude.

Notably, the models featuring structurally regularized latent spaces achieve superior performance, corroborating our \textit{a priori} analysis of their geometric fidelity. The latent Flow Matching model with metric preservation is the top performer, yielding a final ensemble-mean error of only 4.01\%. This constitutes a nearly twofold improvement over the physical-space diffusion model. These results provide strong evidence that explicit geometric regularization of the latent space translates directly to more accurate and stable performance in operational simulations. Across all models, employing an ensemble-mean closure consistently reduces the prediction error by 20-30\% compared to single-sample stochastic trajectories, effectively trading computational cost for reduced variance.

\begin{figure}[!htb]
    \centering
    \includegraphics[width=\linewidth]{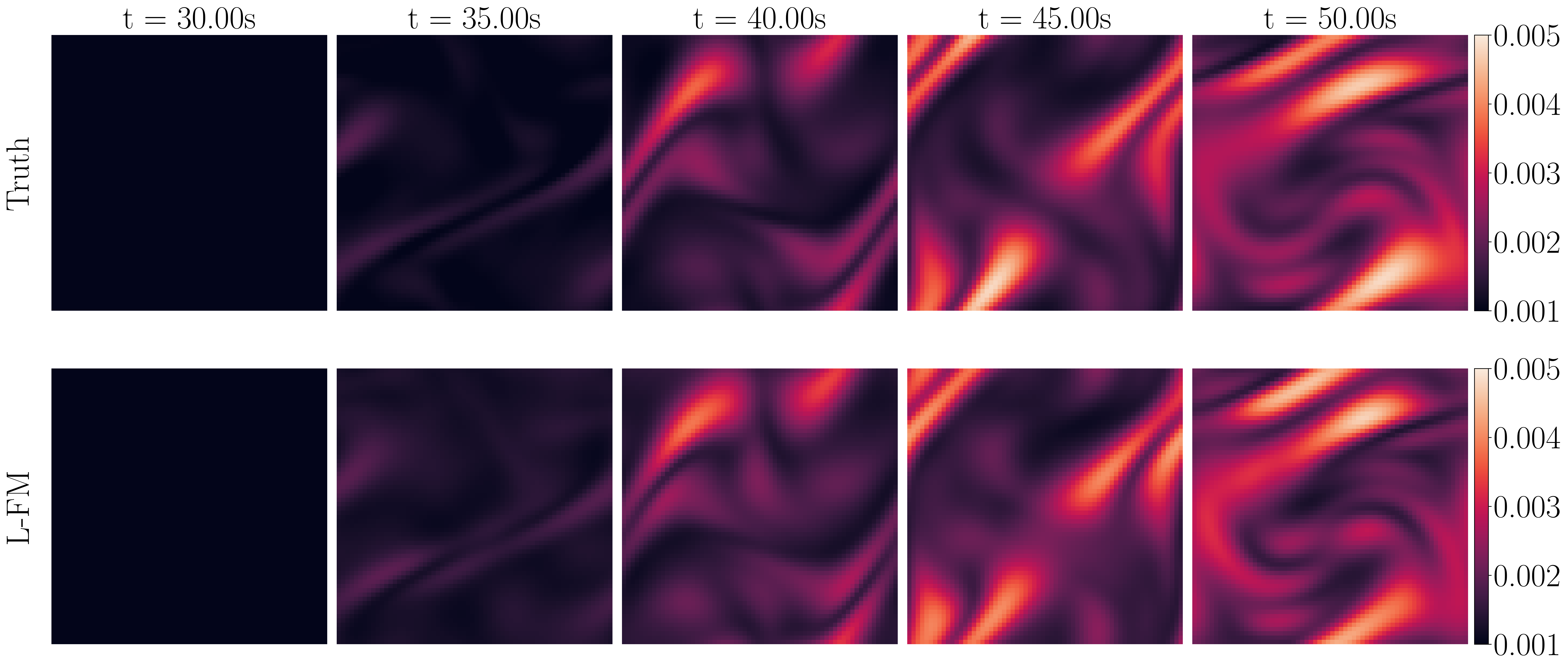}
    \caption{
        Spatial distribution of simulation uncertainty. The pixel-wise standard deviation is computed across an ensemble of 1000 stochastic simulations. The top row shows the ground truth variability, while the bottom row shows the variability captured by the L-FM closure model. The close agreement in both structure and magnitude demonstrates the model's ability to reproduce the physical uncertainty of the system.
    }
    \label{fig: closure_std}
\end{figure}

Beyond accurately predicting the mean flow evolution, a key strength of our stochastic approach is its ability to capture the system's intrinsic variability. Figure~\ref{fig: closure_std} compares the spatial distribution of the standard deviation across an ensemble of stochastic simulations against the ground truth variability. The L-FM model not only predicts the mean state accurately but also reproduces the complex spatial patterns and magnitudes of the system's uncertainty. This demonstrates that the learned conditional distribution $p(H \mid \omega)$ is not merely a source of random noise but a physically meaningful representation of the subgrid-scale dynamics.

\begin{figure}[!htb]
    \centering
    \includegraphics[width=\linewidth]{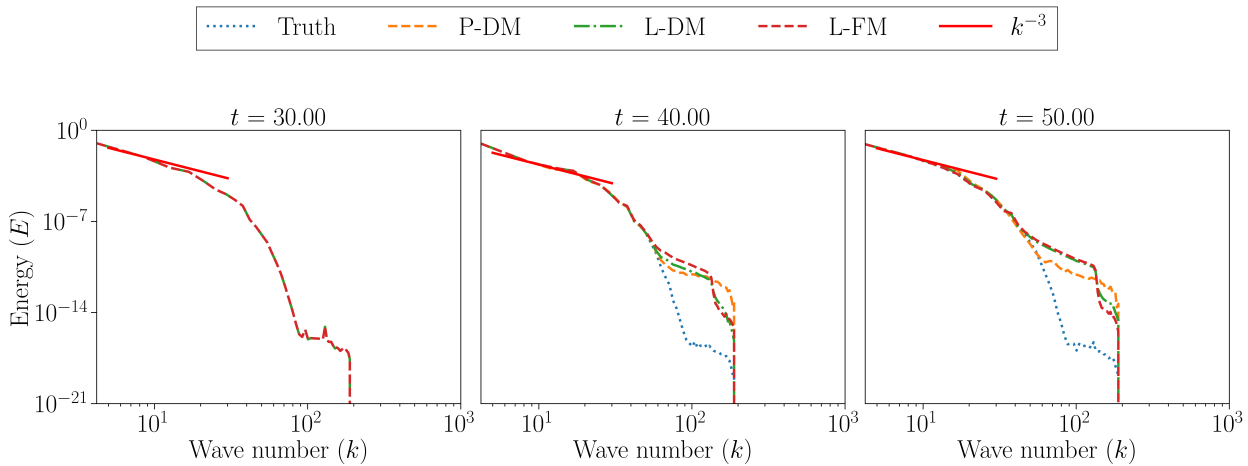}
    \caption{
        Vorticity energy spectra at different time instances. Spectra from simulations using various ensemble-mean closures are compared against the ground truth. All closure-equipped models correctly reproduce the energy distribution and maintain the characteristic $k^{-3}$ slope of the enstrophy cascade, confirming their physical consistency.
    }
    \label{fig: TKE_3050}
\end{figure}

The physical consistency of the closures is further validated by examining the vorticity energy spectra, shown in Figure~\ref{fig: TKE_3050}. All closure-equipped models accurately reproduce the energy distribution of the high-fidelity simulation up to the training resolution limit (wavenumbers $k \approx 10^2$). Crucially, the spectra retain the characteristic $k^{-3}$ power-law decay, indicating that the generative closures correctly preserve the forward enstrophy cascade physics inherent to 2-D turbulence.

Finally, an analysis of computational efficiency highlights the practical advantages of latent-space modeling. As detailed in Table~\ref{tab: simulation_metrics}, generating large ensembles is approximately seven times faster with latent-space models than with their physical-space counterparts due to the reduced dimensionality. Furthermore, the L-FM model, with its one-step generation capability, is twice as fast as the iterative L-DM model while delivering higher accuracy, establishing it as the most effective and efficient model for practical deployment.

\begin{table}[!htb]
\centering
\caption{\textit{A posteriori} simulation performance over a 20-second integration. The relative error ($D_{\mathrm{RE}}$) is reported for both single-trajectory (Per-sample) and ensemble-mean strategies. Per-sample results include the mean and a two-standard-deviation band calculated over 1000 independent trajectories. Computational cost is the wall-clock time per trajectory (Per-sample) or for the full 1000-sample evaluation (Ensemble). All latent models employ MP regularization.}
\begin{adjustbox}{width=0.8\linewidth}
\begin{tabular}{lccccccc}
\toprule
\multirow{2}{*}{\textbf{Model}} 
& \multirow{2}{*}{\textbf{Strategy}} 
& \multirow{2}{*}{\textbf{Cost (s)}} 
& \multicolumn{5}{c}{\textbf{Vorticity field error at time}} \\
\cmidrule(lr){4-8}
& & & \textbf{t=30} & \textbf{t=35} & \textbf{t=40} & \textbf{t=45} & \textbf{t=50} \\
\midrule

No correction & -- & 2.12 & 0 & 4.06e-01 & 6.17e-01 & 6.88e-01 & 8.42e-01 \\
\midrule

\multirow{3}{*}{P-DM} 
& \multirow{2}{*}{Per-sample} & \multirow{2}{*}{180.11} 
& 0 & 3.67e-02 & 5.92e-02 & 1.11e-01 & 1.32e-01 \\
& & & \textcolor{gray}{\scriptsize$\pm$ 0} & \textcolor{gray}{\scriptsize$\pm$ 7.59e-03} & \textcolor{gray}{\scriptsize$\pm$ 9.74e-03} & \textcolor{gray}{\scriptsize$\pm$ 1.04e-02} & \textcolor{gray}{\scriptsize$\pm$ 9.02e-03} \\
\cmidrule(lr){2-8}
& Ensemble & 8662.76 & 0 & 1.66e-02 & 4.13e-02 & 5.02e-02 & 7.55e-02 \\
\midrule

\multirow{3}{*}{L-DM with MP} 
& \multirow{2}{*}{Per-sample} & \multirow{2}{*}{140.66} 
& 0 & 1.96e-02 & 3.95e-02 & 4.17e-02 & 6.72e-02 \\
& & & \textcolor{gray}{\scriptsize$\pm$ 0} &\textcolor{gray}{\scriptsize$\pm$ 1.11e-03} & \textcolor{gray}{\scriptsize$\pm$ 1.20e-03} & \textcolor{gray}{\scriptsize$\pm$ 1.93e-03} & \textcolor{gray}{\scriptsize$\pm$ 2.68e-03} \\
\cmidrule(lr){2-8}
& Ensemble & 1252.49 & 0 & 1.57e-02 & 3.52e-02 & 4.77e-02 & 5.65e-02 \\
\midrule

\multirow{3}{*}{L-FM with MP} 
& \multirow{2}{*}{Per-sample} & \multirow{2}{*}{72.32s} 
& 0 & 1.56e-02 & 3.29e-02 & 3.75e-02 & 4.24e-02 \\
& & & \textcolor{gray}{\scriptsize$\pm$ 0} &\textcolor{gray}{\scriptsize$\pm$ 5.19e-04} & \textcolor{gray}{\scriptsize$\pm$ 1.80e-03} & \textcolor{gray}{\scriptsize$\pm$ 1.80e-03} & \textcolor{gray}{\scriptsize$\pm$ 1.93e-03} \\
\cmidrule(lr){2-8}
& Ensemble & 835.83 & 0 & 1.06e-02 & 2.89e-02 & 3.13e-02 & 4.01e-02 \\
\bottomrule
\end{tabular}
\end{adjustbox}
\label{tab: simulation_metrics}
\end{table}

\section{Conclusion}
In this work, we developed a framework for explicitly regularizing latent diffusion models to build fast and accurate stochastic closures for complex dynamical systems, with the aim of significantly improving the sampling speed when using diffusion models to build stochastic closures of complex dynamical systems. We also systematically compare the performance of explicitly and implicitly regularized latent spaces for several transport-based generative models (diffusion, flow matching, and stochastic interpolants) and find that flow matching is the best-performing sampler due to its straight transport paths, which permit single-step generation. This efficiency is fully realized when paired with a latent space trained with metric-preserving (MP) regularization, while the other implicit regularization (via geometry-aware constraint) or explicit regularization (via joint learning) achieves similar performances for the regularized latent space. The regularized latent space inherits key topological information from the lower-dimensional manifold of the original complex dynamical system and thus enables the use of diffusion models in stochastic closure modeling of high-dimensional complex dynamical systems without demanding a huge amount of training data. When deployed in \textit{a posteriori} simulations of 2D Kolmogorov flow, our framework achieved ten-times faster ensemble simulations, while reducing prediction error by a factor of $O(10)$. We also demonstrated that our framework can provide efficient uncertainty quantification and correctly capture the spatial patterns of the system's intrinsic variability. This work highlights the significant benefits of co-designing machine learning architectures with the underlying geometry of the physical problem, which provides a promising pathway toward extending diffusion-model-based stochastic closures to 3D turbulent flows in science and engineering applications.



\section*{Acknowledgments}
X.D., H.Y., and J.W. are supported by the University of Wisconsin-Madison, Office of the Vice Chancellor for Research and Graduate Education with funding from the Wisconsin Alumni Research Foundation. X.D. and J.W. are also funded by the Office of Naval Research N00014-24-1-2391.

\bibliographystyle{model1-num-names}
\bibliography{references.bib} 

\clearpage

\appendix
\section{Numerical Solver for the 2D Navier-Stokes Equations}\label{app:solver}
The training and evaluation data are generated by solving the 2D incompressible Navier-Stokes system (Eq.~\eqref{eq:2dNS}) using a standard numerical scheme that combines a pseudo-spectral method for spatial discretization with a Crank-Nicolson scheme for time integration.

\subsection{Pseudo-Spectral Method}\label{ssec:pseudospectral}
The pseudo-spectral method is employed for its high accuracy in representing spatially periodic fields. The method leverages the efficiency of the Fast Fourier Transform (FFT) by performing linear operations in the Fourier domain and nonlinear operations in the physical domain.

The simulation is initialized with a vorticity field $\omega(\mathbf{x}, t_0)$ sampled from a statistically stationary Gaussian random field. Given the Fourier coefficients of the vorticity, $\hat{\omega}(\mathbf{k}, t) = \mathcal{F}\{\omega(\mathbf{x}, t)\}$, all linear operations are computed efficiently. The streamfunction $\hat{\psi}$ is found by solving the Poisson equation in the Fourier domain:
\begin{equation}
    \hat{\psi}(\mathbf{k}, t) = \frac{\hat{\omega}(\mathbf{k}, t)}{|\mathbf{k}|^2},
\end{equation}
where $|\mathbf{k}|^2 = k_x^2 + k_y^2$ is the squared wavenumber. The velocity field coefficients $\hat{\mathbf{u}}=(\hat{u}, \hat{v})$ are then derived from the streamfunction:
\begin{equation}
    \hat{u}(\mathbf{k}, t) = i k_y \hat{\psi}(\mathbf{k}, t), \qquad \hat{v}(\mathbf{k}, t) = -i k_x \hat{\psi}(\mathbf{k}, t).
\end{equation}
To compute the nonlinear advection term, $\mathcal{N}(\omega) = -\mathbf{u} \cdot \nabla\omega$, the velocity $\mathbf{u}$ and vorticity gradient $\nabla\omega$ are transformed back to the physical domain, the point-wise product is taken, and the result is transformed back to the Fourier domain. This "pseudo-spectral" approach avoids the expensive convolution operation that a fully spectral method would require.

\subsection{Crank-Nicolson Time Integration}\label{ssec:cranknicolson}
The vorticity equation is advanced in time using a second-order accurate Implicit-Explicit (IMEX) scheme. The stiff linear viscous term is treated implicitly using the Crank-Nicolson method for unconditional stability, while the nonlinear advection and forcing terms are treated explicitly with a forward Euler step. The update rule in the Fourier domain from time $t_n$ to $t_{n+1}$ is:
\begin{equation}
    \frac{\hat{\omega}_{n+1} - \hat{\omega}_n}{\Delta t} = \frac{1}{2}\left(-\nu |\mathbf{k}|^2 \hat{\omega}_{n+1} - \nu |\mathbf{k}|^2 \hat{\omega}_n\right) + \mathcal{F}\{-\mathbf{u}_n \cdot \nabla\omega_n + f + \beta\xi_n\}(\mathbf{k}).
\end{equation}
Rearranging for $\hat{\omega}_{n+1}$ yields the explicit update formula:
\begin{equation}
    \hat{\omega}_{n+1}(\mathbf{k}) = \frac{(1 - \frac{\Delta t}{2} \nu |\mathbf{k}|^2) \hat{\omega}_n(\mathbf{k}) + \Delta t \left( \hat{\mathcal{N}}_n(\mathbf{k}) + \hat{f}(\mathbf{k}) + \beta \hat{\xi}_n(\mathbf{k}) \right)}{1 + \frac{\Delta t}{2} \nu |\mathbf{k}|^2}.
\end{equation}
Recalling from Section~\ref{sec: results} that the closure term is defined as $H = -\mathbf{u} \cdot \nabla\omega + \beta\xi = \mathcal{N}(\omega) + \beta\xi$, we can rewrite the update rule in terms of the closure:
\begin{equation}
    \hat{\omega}_{n+1}(\mathbf{k}) = \frac{(1 - \frac{\Delta t}{2} \nu |\mathbf{k}|^2) \hat{\omega}_n(\mathbf{k}) + \Delta t \left( \hat{H}_n(\mathbf{k}) + \hat{f}(\mathbf{k}) \right)}{1 + \frac{\Delta t}{2} \nu |\mathbf{k}|^2}.
\end{equation}
In forward simulations using a learned stochastic closure, the exact closure term $\hat{H}_n$ is replaced by a sample $\hat{\tilde{H}}_n$ drawn from the generative model at each time step.

\section{Stochastic Forcing via a \texorpdfstring{$Q$}{Q}-Wiener Process}\label{ssec:q_wiener}

The stochastic component $\xi$ in the governing equations (Eq.~\eqref{eq:2dNS}) is modeled as spatially correlated, white-in-time noise. This is formally the time derivative of a $Q$-Wiener process $W(\mathbf{x}, t)$ on a periodic domain $\Omega = [L_1, L_2]^2$. The covariance operator $Q$ is defined in the Fourier basis $\varphi_{\mathbf{k}}(\mathbf{x}) = \exp(i(\lambda_{k_1} x_1 + \lambda_{k_2} x_2))$, where it is diagonal with eigenvalues $q_{\mathbf{k}}$ that prescribe the spatial correlation structure:
\begin{equation}
    Q \varphi_{\mathbf{k}} = q_{\mathbf{k}}\, \varphi_{\mathbf{k}}, \qquad \text{with} \quad q_{\mathbf{k}} = \exp\left(-\alpha(\lambda_{k_1}^2+\lambda_{k_2}^2)\right).
\end{equation}
The parameter $\alpha$ controls the correlation length of the noise.

For numerical implementation on a uniform $N_1 \times N_2$ grid with time step $\Delta t$, a discrete-time realization of the noise field is synthesized from i.i.d.\ complex Gaussian variables $Z_{\mathbf{k}}^{\,n}$. To increase the variance, we aggregate $\kappa$ independent copies. Because $\xi$ is white in time, the variance of its discrete-time realization scales with $1/\Delta t$. The spatially-averaged pointwise variance of the field is given by:
\begin{equation}
    \mathrm{Var}\!\big(\xi^n\big) = \frac{\kappa}{L_1 L_2 \,\Delta t}\, \sum_{\mathbf{k}\in\mathcal{K}_h} q_{\mathbf{k}}.
\end{equation}

The stochastic component of the closure term $H$ is given by $\beta\xi$. We calculate its theoretical standard deviation using the parameters from our numerical setup:
\begin{itemize}
    \item Amplitude: $\beta = 5\times 10^{-5}$
    \item Domain size: $L_1=L_2=1$
    \item Grid size: $N_1=N_2=64$
    \item Time step: $\Delta t = 10^{-3}$
    \item Correlation decay: $\alpha = 5\times 10^{-3}$
    \item Variance inflation factor: $\kappa = 10$
\end{itemize}
First, we numerically compute the sum of the eigenvalues over the discrete grid, which yields $\sum_{\mathbf{k}\in\mathcal{K}_h} q_{\mathbf{k}} \approx 16.0$. The standard deviation of the unscaled noise $\xi^n$ is then:
\begin{align*}
    \mathrm{Std}\!\big(\xi^n\big) &= \sqrt{\frac{\kappa}{L_1 L_2 \,\Delta t}\, \sum_{\mathbf{k}\in\mathcal{K}_h} q_{\mathbf{k}}} \\
    &= \sqrt{\frac{10}{1 \cdot 1 \cdot 10^{-3}} \times 16.0} = \sqrt{16.0 \times 10^{4}} = 400.
\end{align*}
Finally, the standard deviation of the stochastic component of the closure is found by scaling this value by the amplitude $\beta$:
\begin{align*}
    \mathrm{Std}\!\big(\beta\xi^n\big) &= \beta \cdot \mathrm{Std}\!\big(\xi^n\big) \\
    &\approx (5\times 10^{-5}) \times 400 = 2 \times 10^{-2} = 0.02.
\end{align*}

\section{Model Architectures and Training Details} \label{sec:training_details}

Our generative framework consists of two core components: a convolutional autoencoder for dimensionality reduction and a conditional generative model that operates in the latent space.

\subsection{Model Architectures}

The framework is designed to learn a conditional distribution $p(U|V)$, where $U$ and $V$ are high-dimensional fields.

\begin{itemize}
    \item \textbf{Convolutional Autoencoder:} To create an efficient, low-dimensional representation, we use a deep convolutional autoencoder. It maps a high-resolution input field (e.g., $64 \times 64$) to a lower-resolution latent vector (e.g., $16 \times 16$). The encoder and decoder are symmetric, constructed from a series of residual blocks (containing GroupNorm, SiLU activations, and 3$\times$3 convolutions). The encoder uses strided convolutions for downsampling, while the decoder uses transposed convolutions for upsampling. A lightweight self-attention module with four heads is placed at the bottleneck to capture non-local spatial dependencies. We use two identical, independently trained autoencoders: one for the target field $U$ and one for the conditional field $V$.

    \item \textbf{Conditional Generative Model:} The generative model learns a time-dependent vector field $\mathcal{F}_\theta(\tau, z^U_\tau, z^V)$ that defines the transport from a simple prior distribution to the target data distribution in the latent space. Its architecture is based on a Fourier Neural Operator (FNO) and features a two-branch design to process the inputs separately before merging them.
    \begin{enumerate}
        \item \textbf{Target Branch:} The intermidiate time-dependent latent state $z^U_\tau$ is processed. First, the transport time $\tau \in [0,1]$ is encoded into a vector using sinusoidal Gaussian Fourier features and a small MLP. This time embedding is then concatenated with $z^U_\tau$ and a set of normalized spatial coordinates $(x,y)$.
        \item \textbf{Conditional Branch:} The conditional latent $z^V$ is concatenated with the same spatial coordinates.
    \end{enumerate}
    Each branch consists of four Fourier layers, which apply convolutions in both the spatial and frequency domains, interleaved with GELU activations. The outputs of the two branches are then concatenated channel-wise and fused using a final 1$\times$1 convolutional network to produce the vector field estimate (e.g., score function in diffusion models, velocity field in flow matching and stochastic interpolants).
\end{itemize}

\subsection{Training Protocol}

Our training data is sourced from 100 high-fidelity simulations of the 2D Navier-Stokes equations, from which we extract 20,000 paired snapshots of the resolved vorticity $\omega$ and the closure term $H$ over a 20-second interval. This dataset is split into training (18,000), validation (1,000), and test (1,000) sets. We investigate two primary training strategies:

\begin{itemize}
    \item \textbf{Two-Phase Training:} This is a sequential approach where the autoencoders and generative model are trained separately.
    \begin{enumerate}
        \item \textit{Phase 1: Autoencoder Training.} The autoencoders are trained on the 18,000 full-resolution snapshots. For the conventional two-phase model, the training objective is solely the mean squared reconstruction error (MSE). For the structurally regularized models, this MSE loss is augmented with either the Metric-Preserving (MP) or Geometry-Aware (GA) loss term. We use the Adam optimizer ($lr=10^{-3}$), a batch size of 200, a `ReduceLROnPlateau` scheduler, and early stopping to find the optimal weights.
        \item \textit{Phase 2: Latent Model Training.} After freezing the optimal autoencoder weights, we encode the entire training set to get 18,000 latent pairs $(z^\omega, z^H)$. The FNO-based generative model is then trained on these pairs for 1000 epochs using Adam ($lr=10^{-3}$) with a step-decay schedule and a batch size of 200.
    \end{enumerate}

    \item \textbf{End-to-End Joint Training:} In this strategy, the autoencoders and the generative model are optimized simultaneously. The total loss is a weighted sum of the autoencoder reconstruction losses, the generative model's transport loss, and a KL-divergence term on $z^H$ to regularize the latent space. Based on a grid search, we set the loss weights to $\lambda_H=10$, $\lambda_\omega=0.1$, $\lambda_{\mathrm{transport}}=0.1$, and $\lambda_{\mathrm{KL}}=0.01$. The optimizer and scheduler settings are identical to those used in the AE training phase.
\end{itemize}
\end{document}